\documentclass[11pt]{article} 

\usepackage[utf8]{inputenc} 

\usepackage{geometry} 
\usepackage{graphicx} 
\usepackage{amsmath}

\usepackage{booktabs} 
\usepackage{array} 
\usepackage{paralist} 
\usepackage{verbatim} 
\usepackage{subfig} 

\usepackage{fancyhdr} 
\usepackage{sectsty}
\allsectionsfont{\sffamily\mdseries\upshape} 

\usepackage[nottoc,notlof,notlot]{tocbibind} 
\usepackage[titles,subfigure]{tocloft} 

\title{Color Image Edge Detection using Multi-scale and Multi-directional Gabor filter}

\author{Yunhong Li , Yuandong Bi , Weichuan Zhang, Jie Ren and Jinni Chen}

\date{} 

\begin{document}
\maketitle

\abstract{In this paper, a color edge detection method is proposed where the multi-scale Gabor filter are used to obtain edges from input color images. The main advantage of the proposed method is that high edge detection accuracy is attained while maintaining good noise robustness. The proposed method consists of three aspects: First, the RGB color image is converted to CIE L*a*b* space because of its wide coloring area and uniform color distribution. Second, a set of Gabor filters are used to smooth the input images and the color edge strength maps are extracted, which are fused into a new ESM with the noise robustness and accurate edge extraction. Third, Embedding the fused ESM in the route of the Canny detector yields a noise-robust color edge detector. The results show that the proposed detector has the better experience in detection accuracy and noise-robustness.}

\section *{Keywords: Edge detection; Gabor filter; Canny Detector; Multi-scale multiplication technique; CIE L*a*b*.}

\section{Introduction}
\label{sec:introduction}
Edge detection~\cite{JING2022259} is a foundational operation in computer vision and image processing. Edges refer to positions where the information of pixels change suddenly in gray-scale or color images~\cite{Torre,upla2015an,shui2012noise-robust,ZHANG2017193, Yunhong2020,jing2021novel}. Many image processing and analysis techniques make use of edge information, such as image segmentation~\cite{shi2000normalized,comaniciu2002mean,Rivera}, image retrieval~\cite{manjunath1996texture,cheng2015global,Zhang2013}, and corner detection~\cite{6507646,Zhang2019,Zhang2014IET,ZHANG20152785}. The edge detection result of color image is $10\%$ of total edge pixels more than that of gray-scale image, and these $10\%$ color edge pixels usually play a key role in certain applications~\cite{lupyan2015object,wang2016noise-robust}. Furthermore human perception of color images is more richer than the achromatic pictures~\cite{masia2013special,1977Nevatia}. Therefore, an efficient and reliable edge detection algorithm in color images is very important for many computer vision tasks.

The numerous edge detection methods have been developed for four decades and can be roughly divided into three groups~\cite{ruzon2001edge}: output fusion methods, gradient-based methods and vector methods. In output fusion methods, the final edge maps are yielded by combining the independent edge detections from each color component~\cite{topal2012edge}. Estimating and calculating the orientation and strength of each edge pixel is called multidimensional gradient methods~\cite{khotanzad1989unsupervised}. For the vector methods, the vector nature of the color image can be kept throughout computation. As a generalization of the gray-scale morphological gradient, the color morphological gradient(CMG) is used in color edge detection~\cite{evans2006a}, which avoids fusion operation. This paper focuses on the multidimensional gradient methods.

The multidimensional gradient methods are characterized by looking for the orientations and strengths of the edges, and fusion operation is before bi-threshold decision. In 1977, One of the earliest color edge detectors is proposed by Navatia, who transformed the image data to luminance and two chrominance components and used Huckel's edge detector to obtain the edge map in each individual component independently~\cite{1977Nevatia}. But the shortcoming is the simple orientation.  Robinson~\cite{robinson1977color}, who also proposed a method on color edge detection in the same year. The 24 directional derivatives are computed and the one of the largest magnitude is chosen as the gradient. Tsang and Tsang proposed a novel color edge detection algorithm based on extraction of gradient discontinuities in the HSV color space~\cite{tsang1997suppression}, which is capable of suppressing false edge detection in specular reflective regions.  An edge detection of color images using local directional operators is provided by Scharcanski and Venetsanopoulos~\cite{scharcanski1997edge}. The edge direction information is a relevant feature to a variety of image analysis tasks~\cite{Tai2008, Deng2011,jing2021image, zhang2020corner, 8883063}. Recently, the Edge drawing-based detector(colorED) that combines the gradient with smart routing is proposed for color edge detection~\cite{akinlar2017colored}.

The Canny edge detector~\cite{canny1986a}, as a milestone, has been developed for thirty years and have been improved as many detectors. The flowchart of Canny algorithm mainly has three steps: image smoothing, non-maximum suppression and hysteresis threshold. After the aforementioned operations, the candidate edge pixels should be found and classified as strong edges and weak edges. Then some weak edges should be filled out as edges owing to the continuity of edges. The output edge images are attained.  Kanade introduced the Canny edge operator for color edge detection in ~\cite{Kanade}.The extraction of the magnitude and direction of the edge is the key to obtain high-quality edges. Research shows that the Gabor filter is similar to  perception of the human visual system~\cite{1985Uncertainty,hill2016contrast,ni2018a}. Liu and Wechsler pointed out the merits of Gabor filters in extracting local features of the images~\cite{liu2002gabor}. Zhang used Gabor filters to smooth edges and detect corners using amplitude response and angle response~\cite{Zhang2014IET}. A set of multi-scale and multi-directional Gabor filters are used in the proposed algorithm to detect the corners of the images. In this paper, the RGB color image is converted to CIE L*a*b space, which is composed of one luminance channel and two chromaticity channels and is sensitive to human visual perception. Then the multi-scale Gabor filters are used to smooth the images for the edge strength maps(ESMs). The Gabor filters with a small scale is wit high edge resolution, edge localization while it is sensitive to noise. On the contrary, the Gabor filters with a large scale is noise robustness while it is worse in edge accuracy. The fused ESM from the edge strength maps(ESMs) is embedded into the framework of Canny detection for obtaining edge contours. The proposed edge detector is compared with Color Canny, Laplacian, CMG, improved Sobel, AGDD and ColorED methods~\cite{Kanade,Tai2008,evans2006a,Deng2011,akinlar2017colored,wang2016noise-robust}. The input images from the BSD500 dataset are assessed the performance of the accuracy of the detectors. And the performance of the noise robustness are assessed by FOM measure. The experimental results show that the proposed method is of very high quality.

This paper is organized as follows. Section \uppercase\expandafter{\romannumeral2} first introduces the conversion of RGB space to CIE L*a*b* space. Then the multi-scale Gabor filters are introduced. In Section \uppercase\expandafter{\romannumeral3}, the color edge detector using the Gabor filters is proposed. The multi-scale edge strength maps of the Gabor filters are presented. In Section \uppercase\expandafter{\romannumeral4},  a new edge detection measure is derived which has good edge detection accuracy, good edge localization, and noise robustness.  A full performance evaluation of the proposed method is reported in Section \uppercase\expandafter{\romannumeral5}. Finally, conclusion is given.

\section{Related work}
\label{sec:1}
In this section, the conversion of RGB space to CIE L*a*b* space is introduced first. Then, the multi-scale and directional Gabor filters are presented.

\subsection{The conversion of RGB space to CIE L*a*b* space}
\label{sec:2}

\begin{figure}[t!]
\begin{center}
\begin{tabular}{c}
\includegraphics[width=15cm, height=15cm]{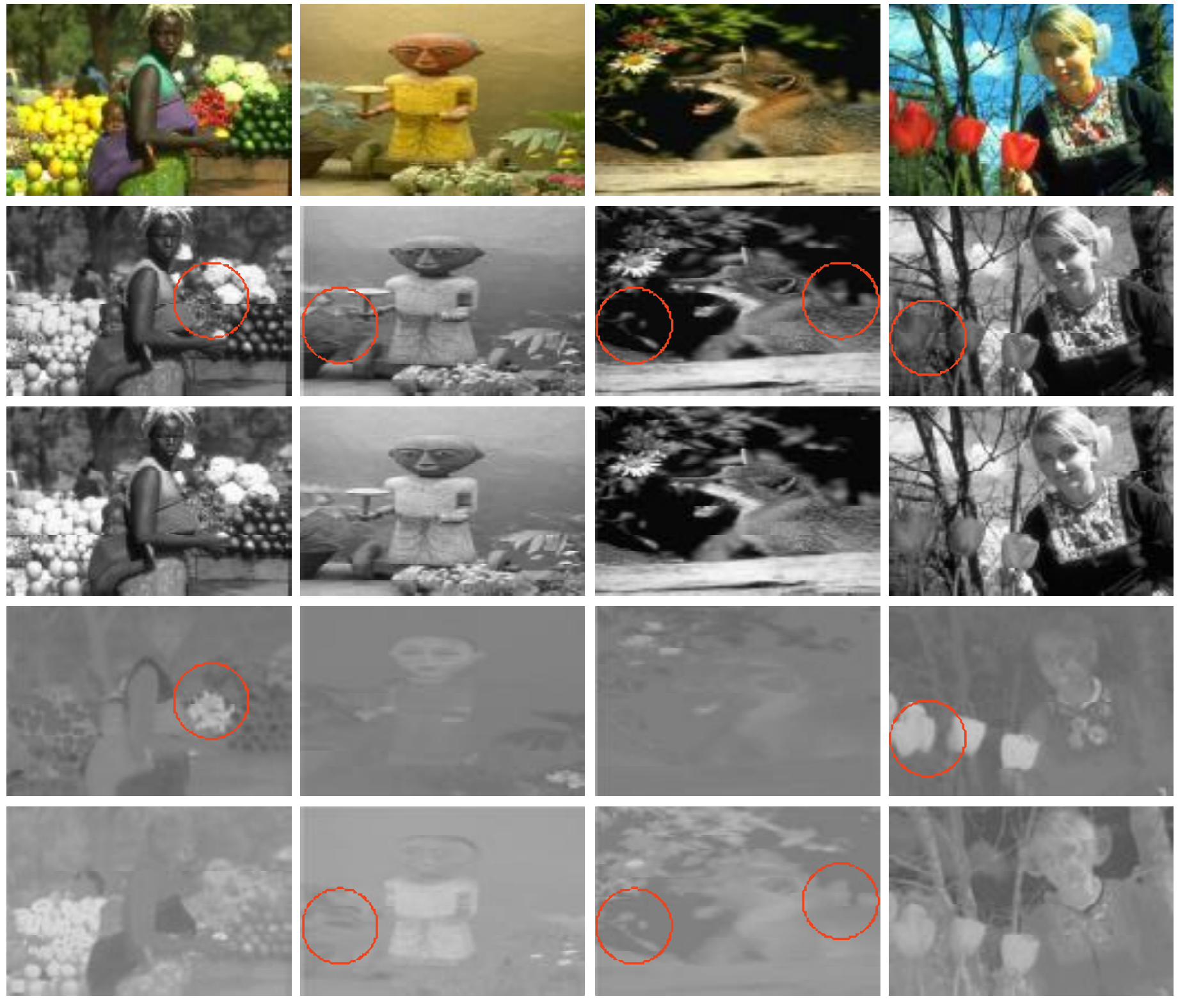}
\end{tabular}
\end{center}
\caption{Four original color images and the corresponding images in different channels.(From top to the bottom, the gray-scale images and the images in L*, a* and b* channels are shown in the first row to the fifth row, respectively.)}
\label{f0}
\end{figure}

The CIEL*a*b* color space was published by the International Commission on Illumination (CIE) in 1976. This color mode consists of three channels: L is the luminance channel, a* and b* are chromaticity channels. And value ranges of L is in [0, 100], the range of a* and b* values are both in [-128, 127]. The CIE L*a*b* space  can better reflect the color difference of objects. Therefore, it has been shown powerful representation ability of visual information and widely used in various image analysis tasks~\cite{jing2016the,lei2017a}.

The RGB color space cannot be directly converted to CIEL*a*b color space. The CIE XYZ~\cite{liang2013salient} color space is required as a medium. The conversion of the CIE L*a*b* space to the CIE XYZ is
\begin{equation}
\begin{bmatrix}
X \\
Y \\
Z
\end{bmatrix}
=
\begin{bmatrix}
0.4124 & 0.3575 & 0.1804 \\
0.2128 & 0.7152 & 0.0722 \\
0.0193 & 0.1192 & 0.9502
\end{bmatrix}
\begin{bmatrix}
R      \\
G      \\
B
\end{bmatrix}.
\label{eq1}
\end{equation}

The conversion of the CIE L*a*b space to the CIE-XYZ is

\begin{equation*}
L^*=116f(Y/Y_0)-16,
\end{equation*}

\begin{equation*}
a^*=500[f(X/X_0)-f(Y/Y_0)],
\end{equation*}

\begin{equation}
b^*=200[f(Y/Y_0)-f(Z/Z_0)],
\label{eq2}
\end{equation}

\begin{equation}
f(t)=
\begin{cases}
t^{1/3},  & \mbox{if }t>0.008856 \\
7.787t+\frac{4}{29}, & \mbox{otherwise}
\end{cases},
\label{eq3}
\end{equation}
where $X_0$, $Y_0$ and $Z_0$ are the tristimulus value of the CIE standard illuminator. $X_0=95.047$, $Y_0=100$ and $Z_0=108.883$.

In Fig ~\ref{f0}, original color images, the corresponding gray-scale images and the images in L*, a* and b* channels are shown in the first row to the fifth row, respectively. In the picture of "Women and child" in the first column on the left, the area of fruit marked by the red circle is not easy to be distinguished in the gray-scale image. But it is easy in the corresponding image in the a* channel because of the difference in chromaticity. As The region marked by the red circle in the gray-scale picture of "Pottery figurine" shown, the difference between the plant and the back wall in the corresponding gray-scale image is small, which causes it to be covered by the background. However, its overall shape and structure are revealed in the b* channel of the image. The lines of branches are discontinuous and fuzzy in the gray-scale picture of "Puppy" while these lines are complete and the outline is clear in b* channel of the image. The roses are overlapped with the background in the region marked by the circle in the picture of "Girl". Due to the advantage of chromaticity, the lines and outlines in the a* channel of the image are displayed more completely than they in the gray-scale image. Therefore, it can be concluded that using the CIE L*a*b* color space is very beneficial for extract the detailed information of the edges.

\begin{figure}
\begin{center}
\begin{tabular}{cccc}
\includegraphics[width=3cm, height=4cm]{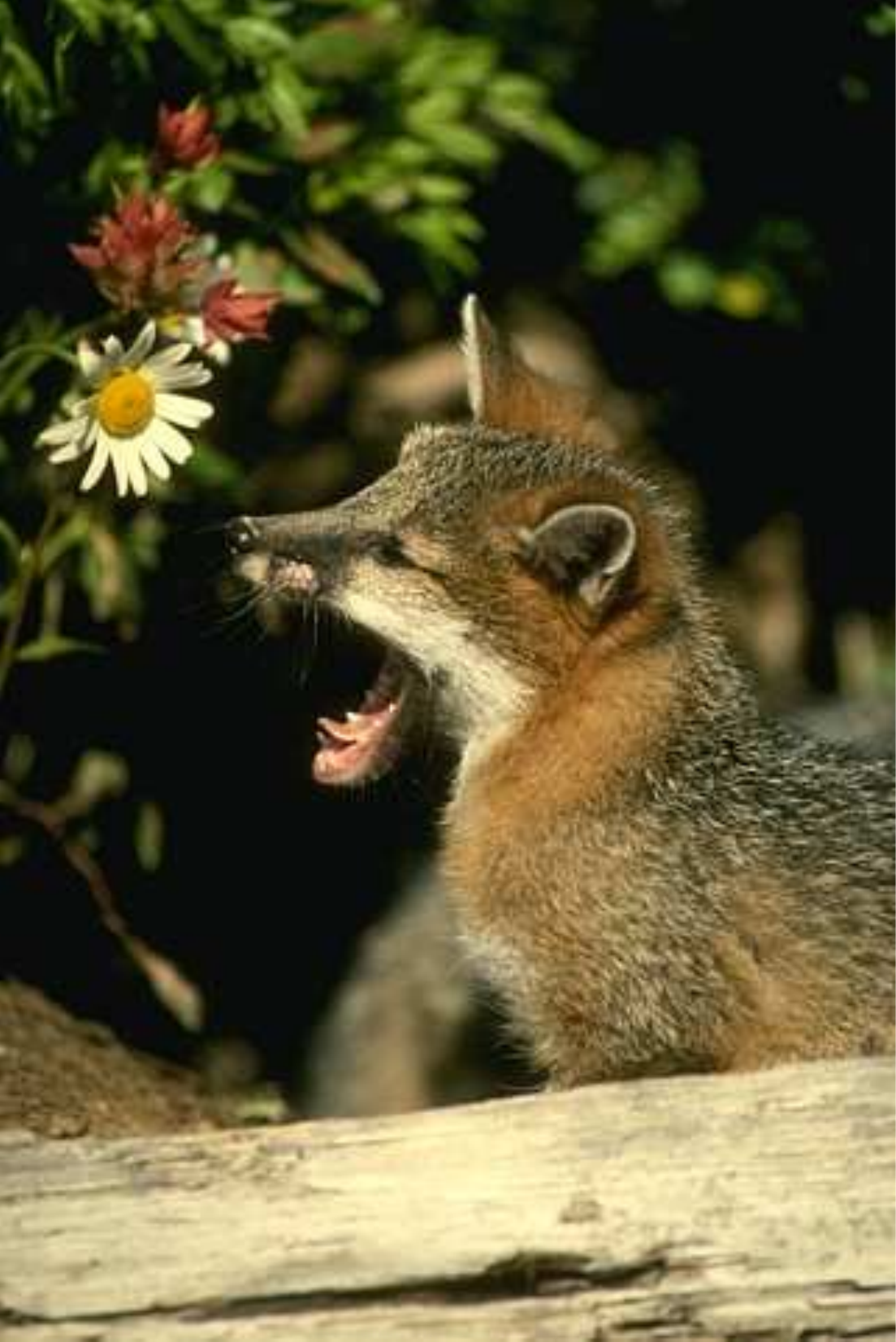}\label{1a}&
\includegraphics[width=3cm, height=4cm]{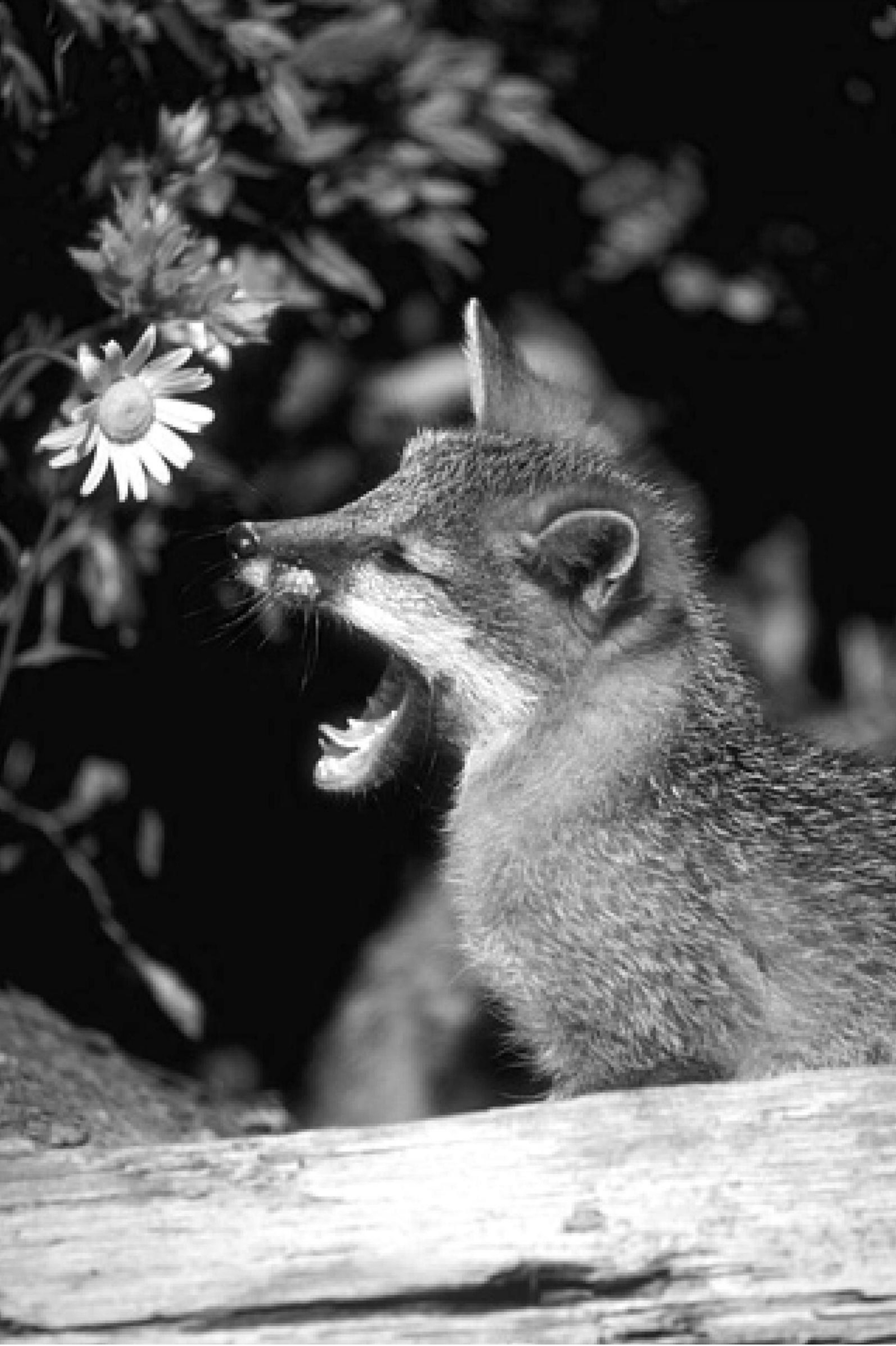} \label{1b}&
\includegraphics[width=3cm, height=4cm]{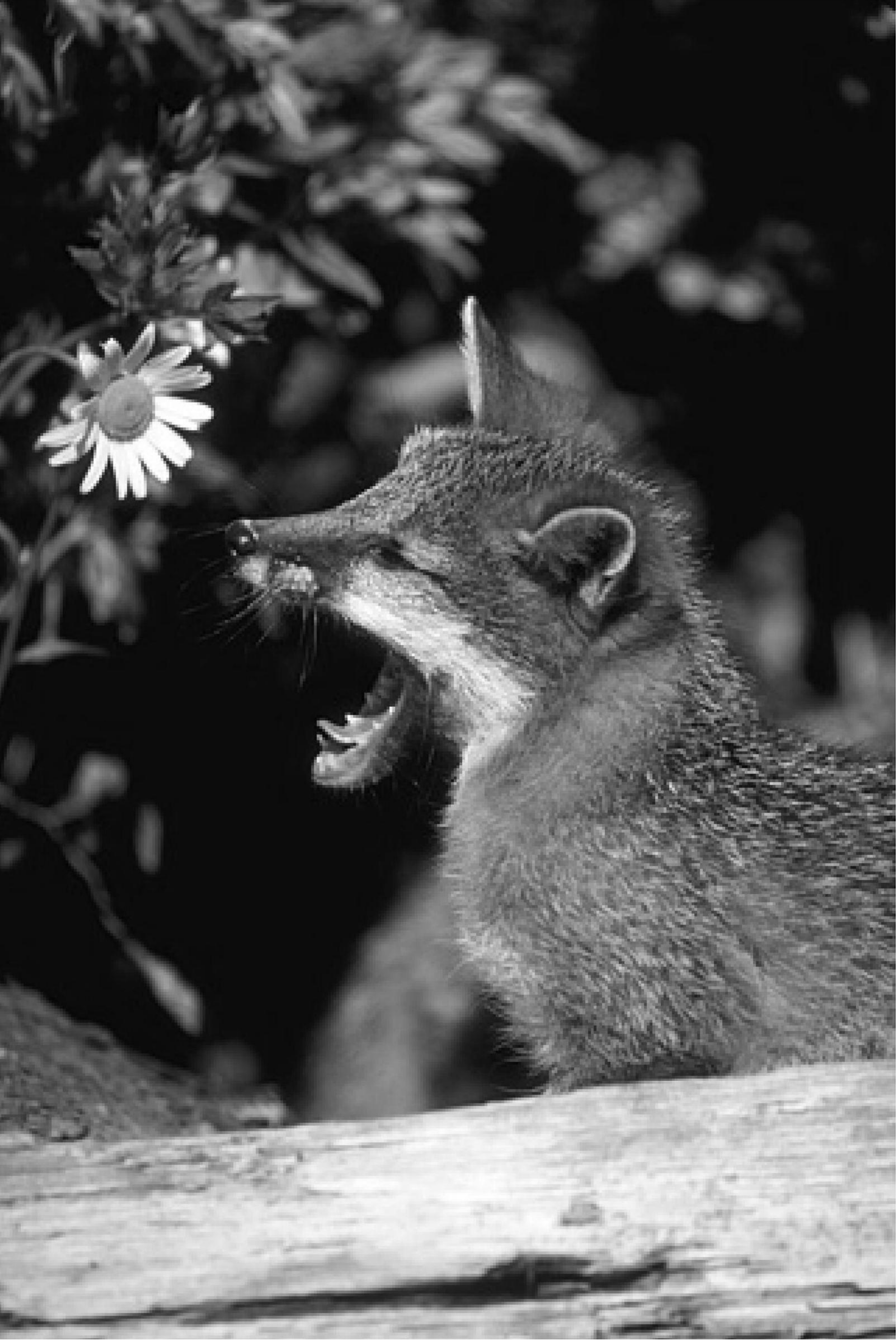} \label{1c}&
\includegraphics[width=3cm, height=4cm]{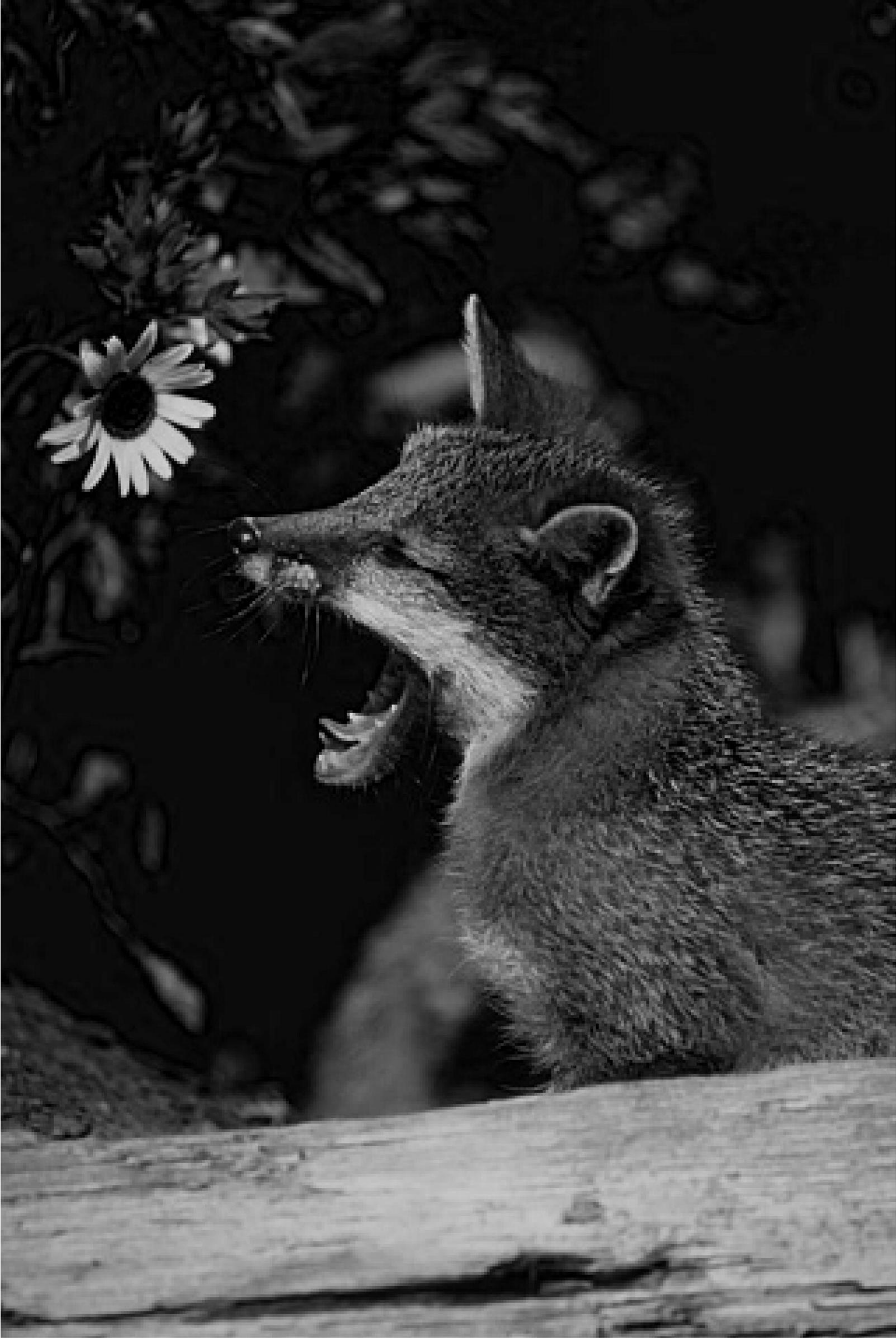} \label{1d}
\\
(a) & (b) & (c) & (d)\\
                                                                                                 &
\includegraphics[width=3cm, height=4cm]{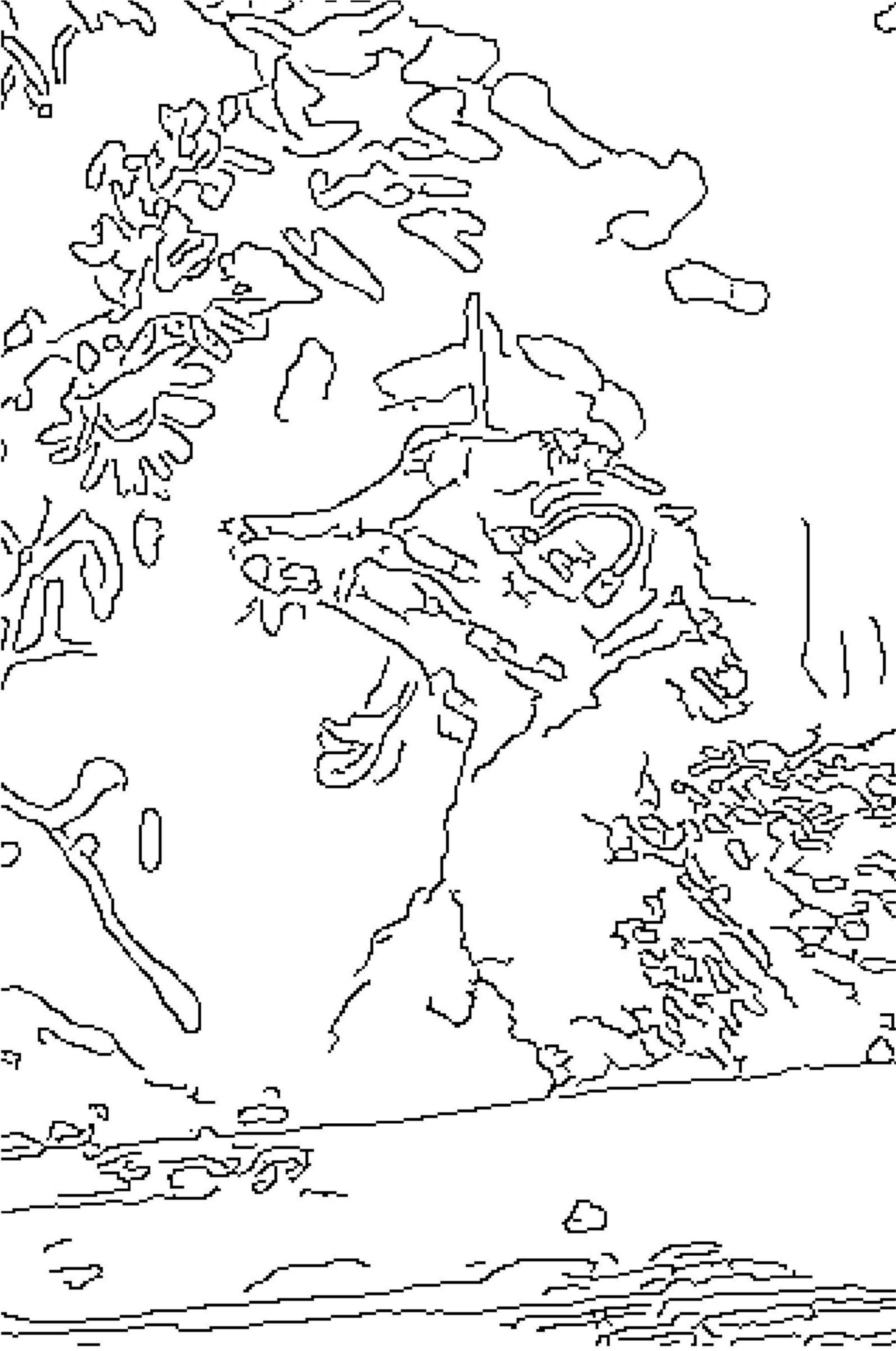} \label{1e}&
\includegraphics[width=3cm, height=4cm]{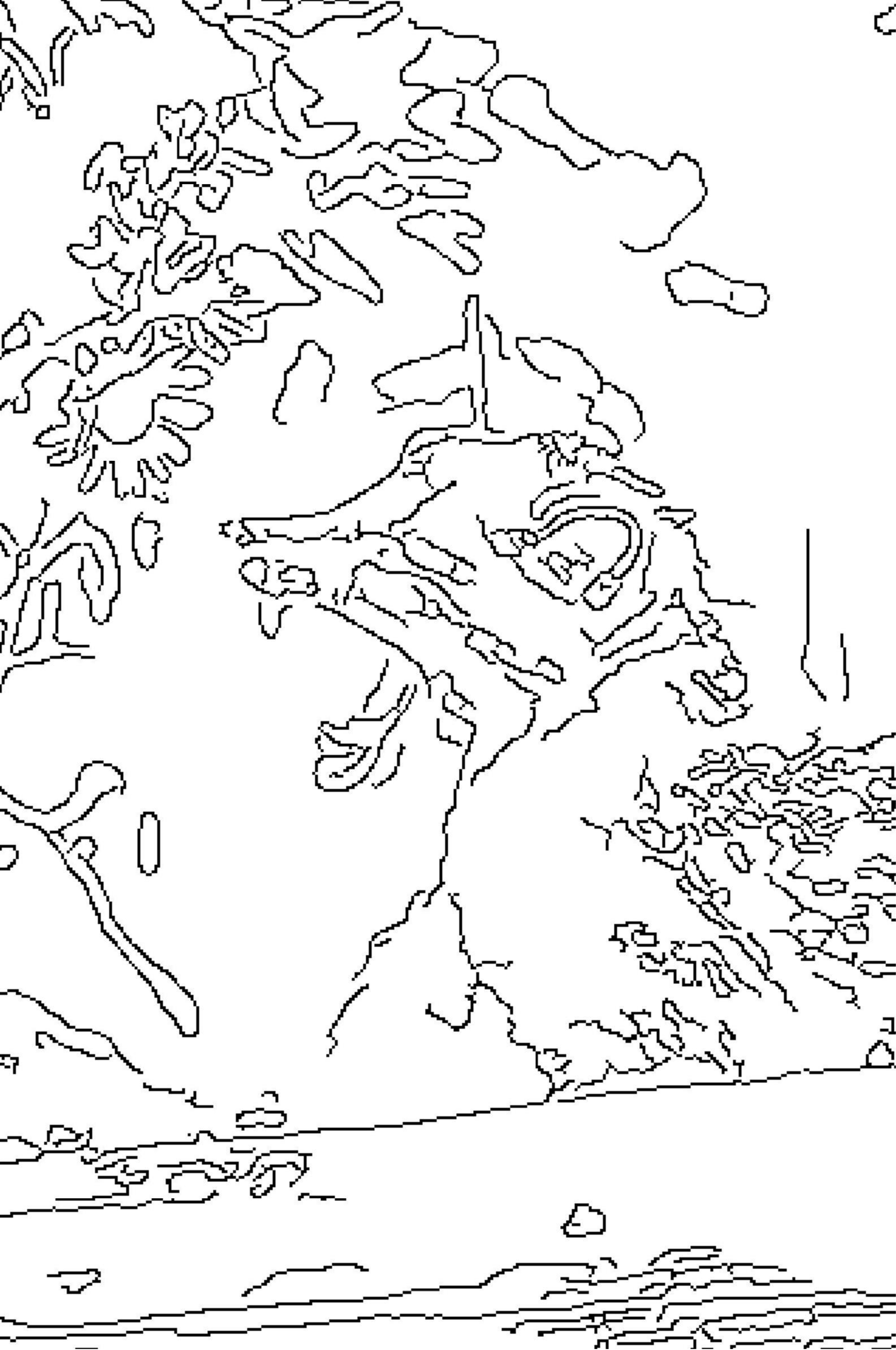} \label{1f}&
\includegraphics[width=3cm, height=4cm]{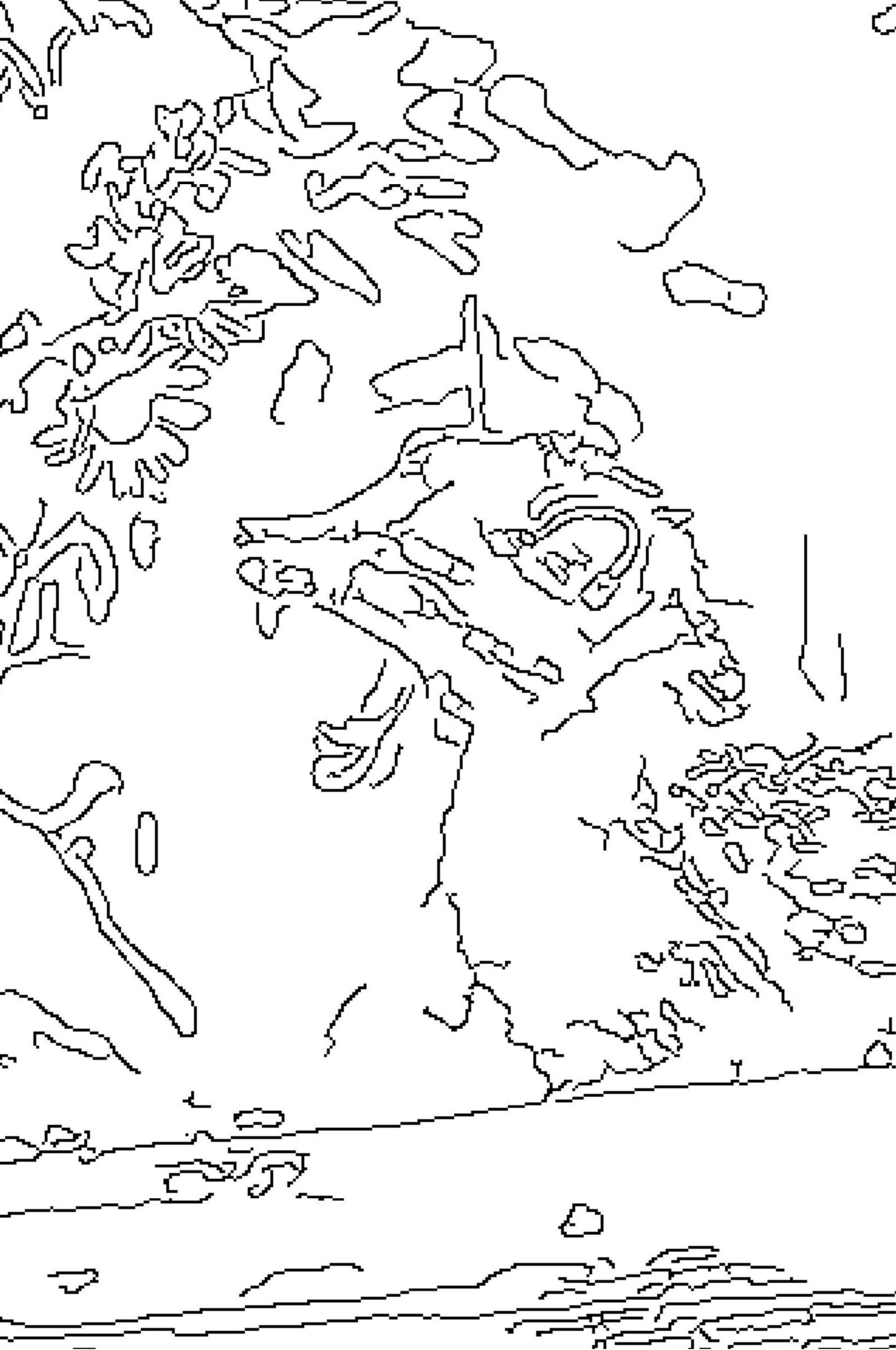} \label{1g}
\\
      & (e) & (f) & (g)\\
                                                                                                 &
\includegraphics[width=3cm, height=4cm]{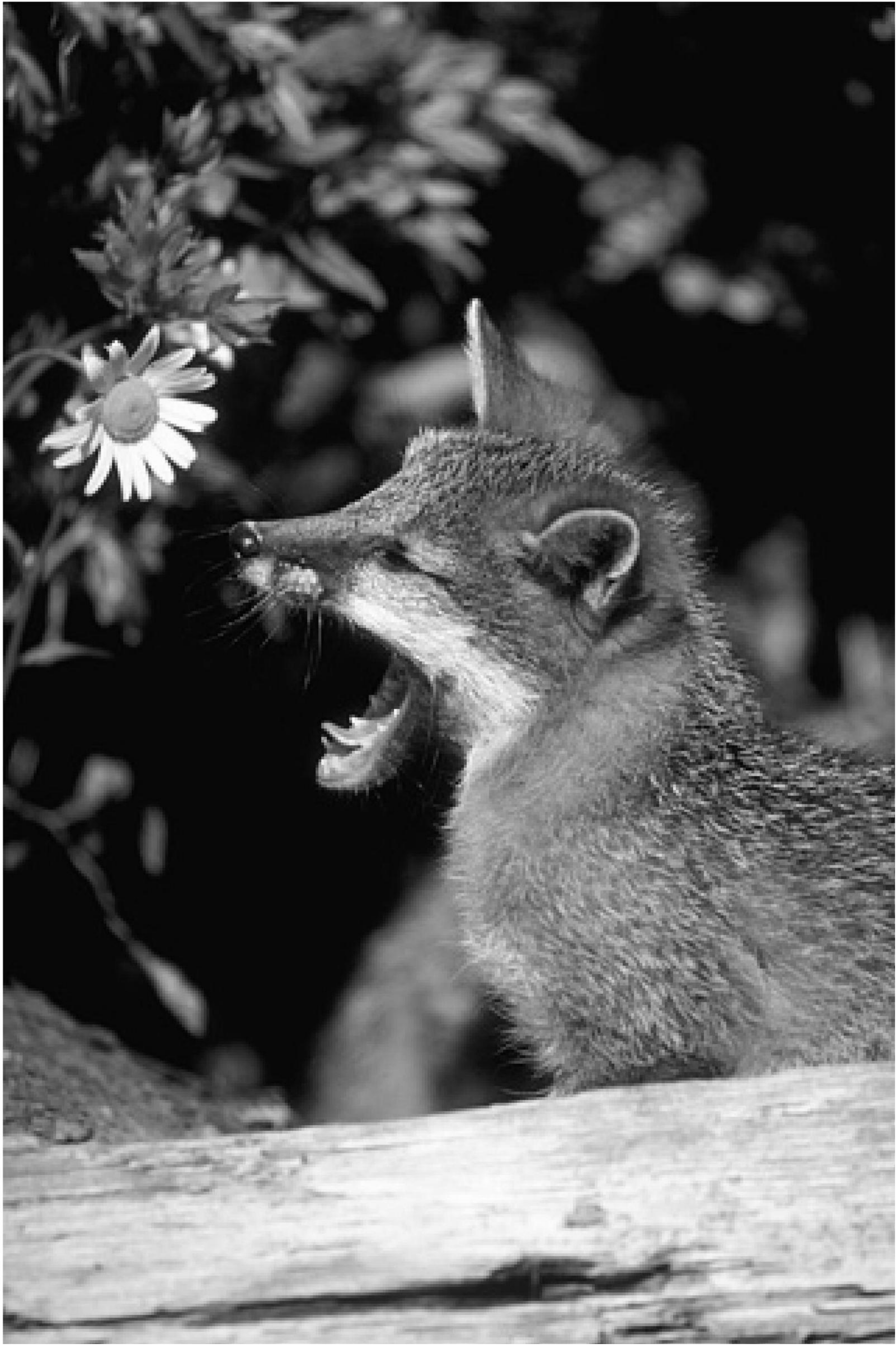} \label{1h}&
\includegraphics[width=3cm, height=4cm]{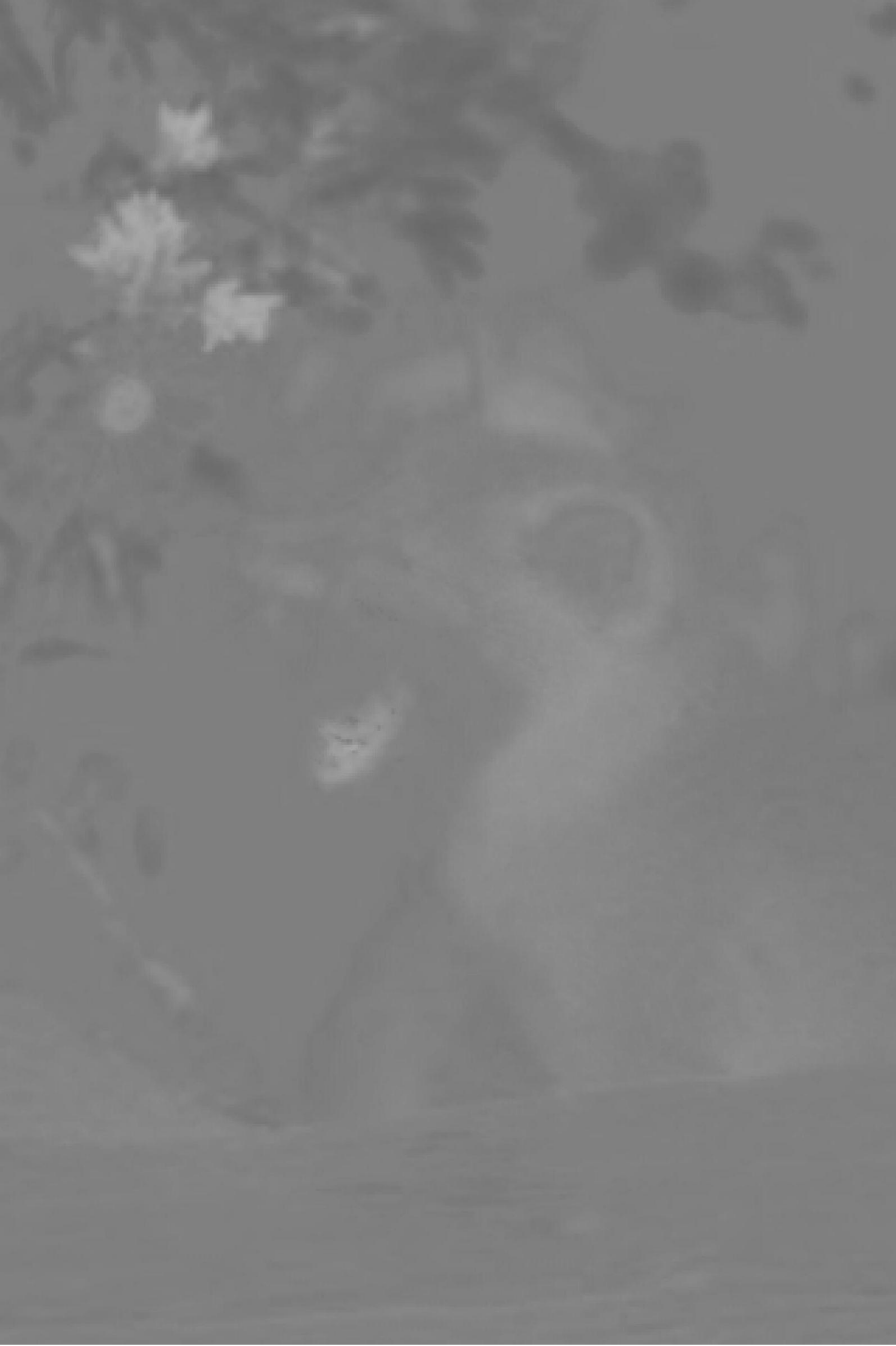} \label{1i}&
\includegraphics[width=3cm, height=4cm]{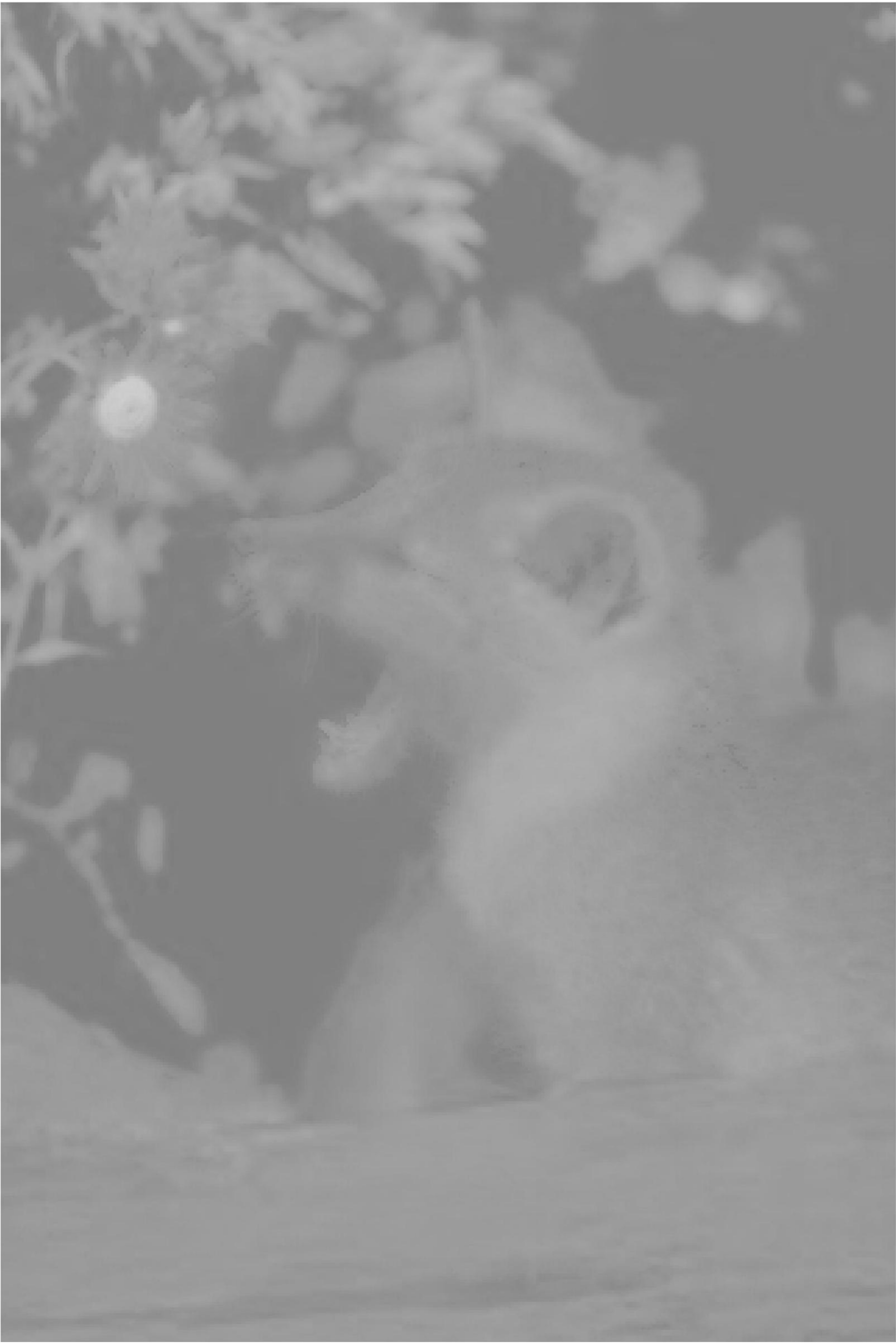} \label{1j}
\\
      & (h) & (i) & (j)\\
                                                                                                   &
\includegraphics[width=3cm, height=4cm]{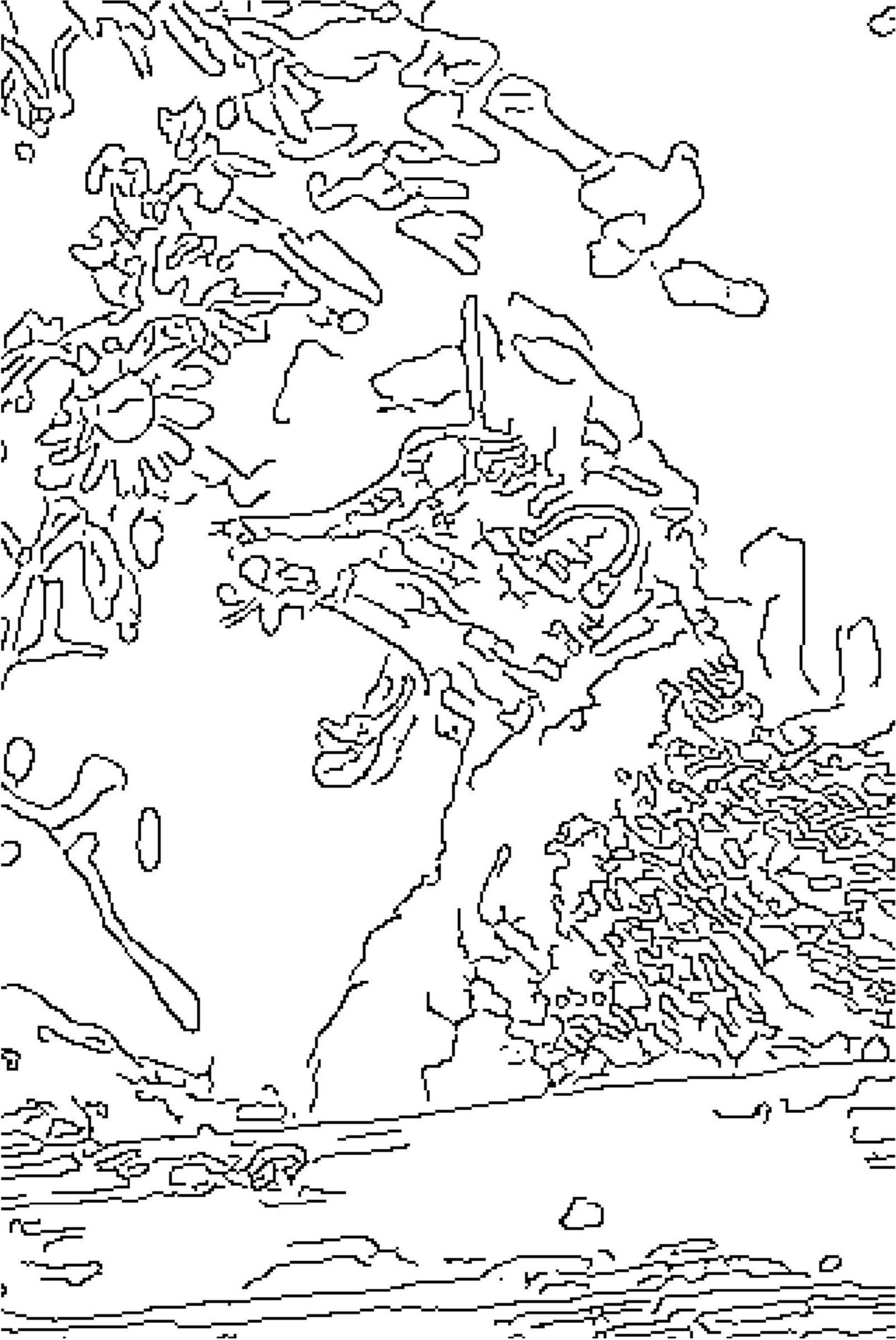} \label{1k}&
\includegraphics[width=3cm, height=4cm]{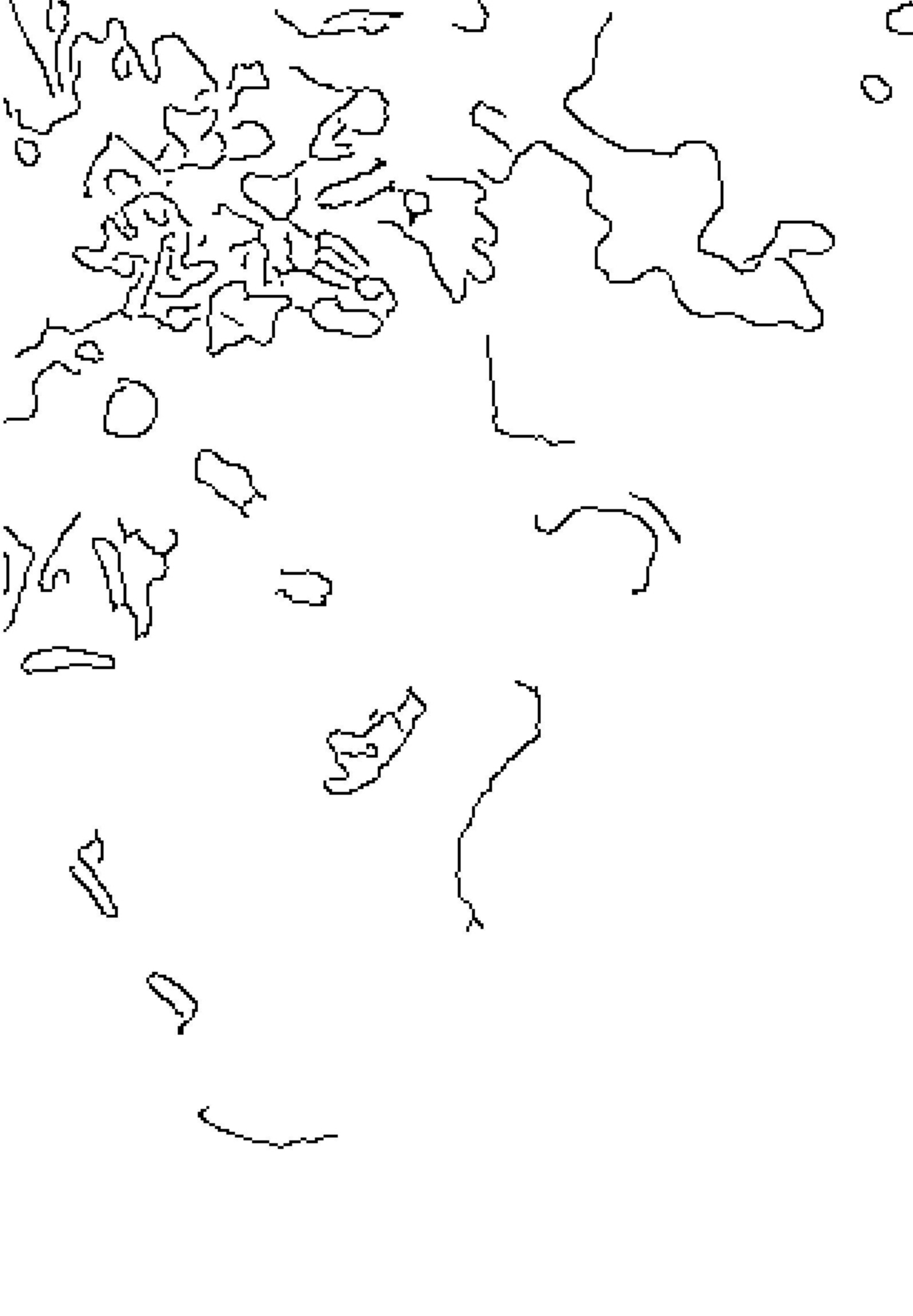} \label{1l}&
\includegraphics[width=3cm, height=4cm]{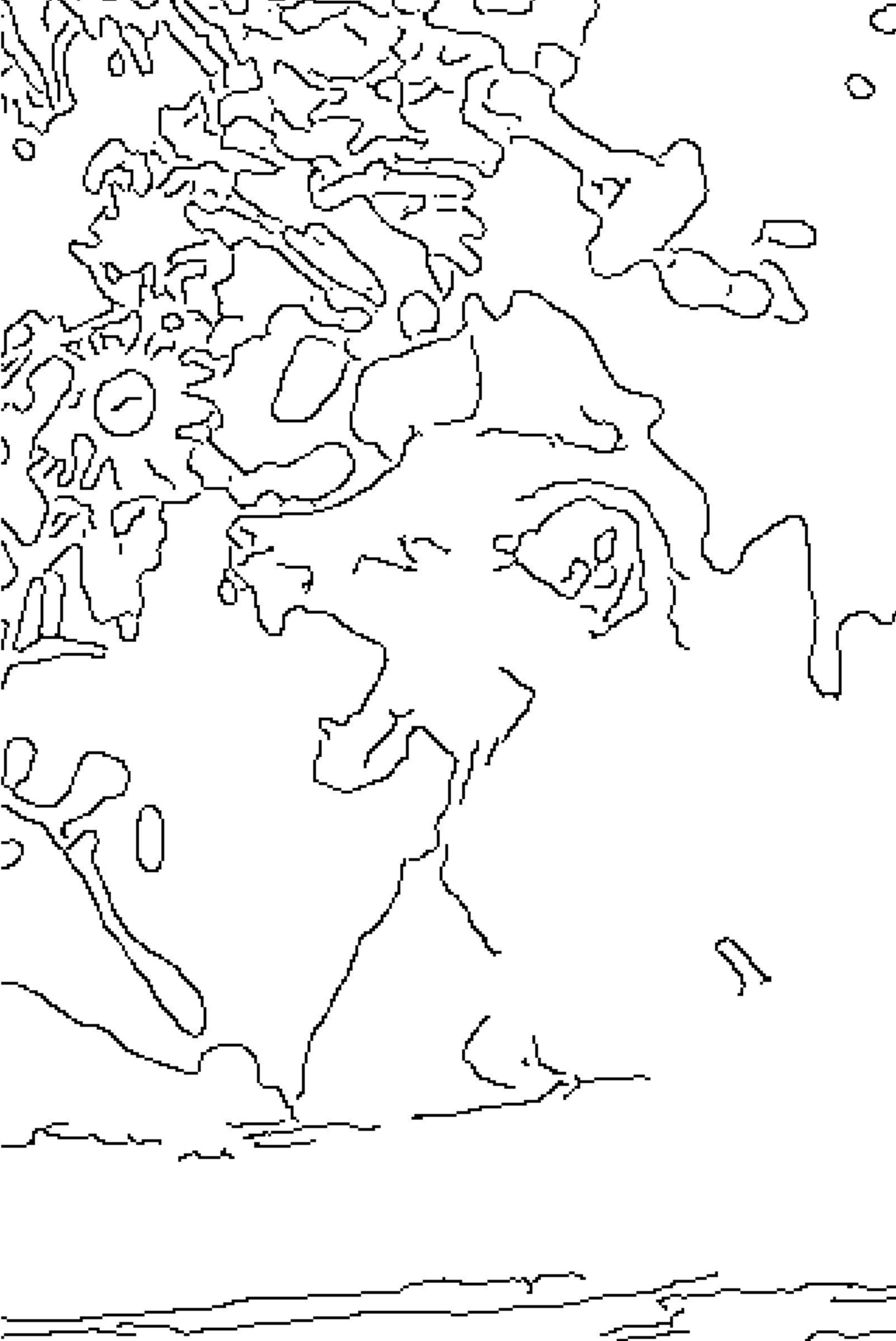} \label{1m}
\\
      & (k) & (l) & (m)\\
\end{tabular}
\end{center}
\caption{The "Puppy" images in R, G, B, L*, a* and b* channels and their corresponding detection results: (a) an original image; (b) R channel of the original image; (c) G channel of original the image; (d) B channel of original the image; (e), (f) and (g) are the detected results of R, G and B channels of original image, respectively ; (h) L* channel of the original image; (i) a* channel of original the image; (j) b* channel of original the image; (k), (l) and (m) are the detected results of L*, a* and b* channels of original image, respectively.}
\label{f1}
\end{figure}

Comparing with RGB color space, CIE L*a*b* color space has more obvious edges information in Fig ~\ref{f1}. It can be observed from Fig ~\ref{f1} (e), (f) and (g) that the relatively complete edge detection results can be attained by the R, G, and B channels of the images. However, the detected results fail to show large differences, which leads to the loss of some edge details. The result of edge detection is incomplete in the position of the two red circles in the "Puppy" picture in Fig ~\ref{f0}. As the L*, a*, and b* channels of the images and their corresponding detected results in the Fig ~\ref{f1} (h), (i), (j), (k), (l) and (m) shown, although the detected result of the image in the a* channel is not ideal and the missing is more serious, the large differences can be observed in L* and b* channels. And the Overlay operation is used later. Obviously, in the region marked by two red circles in the "Puppy" , the result of edge detection is relatively continuous and complete. not only the display of different brightness levels, but also the large differences in chromaticity.

\begin{figure}
\begin{center}
\begin{tabular}{cc}
\includegraphics[width=7cm, height=6cm]{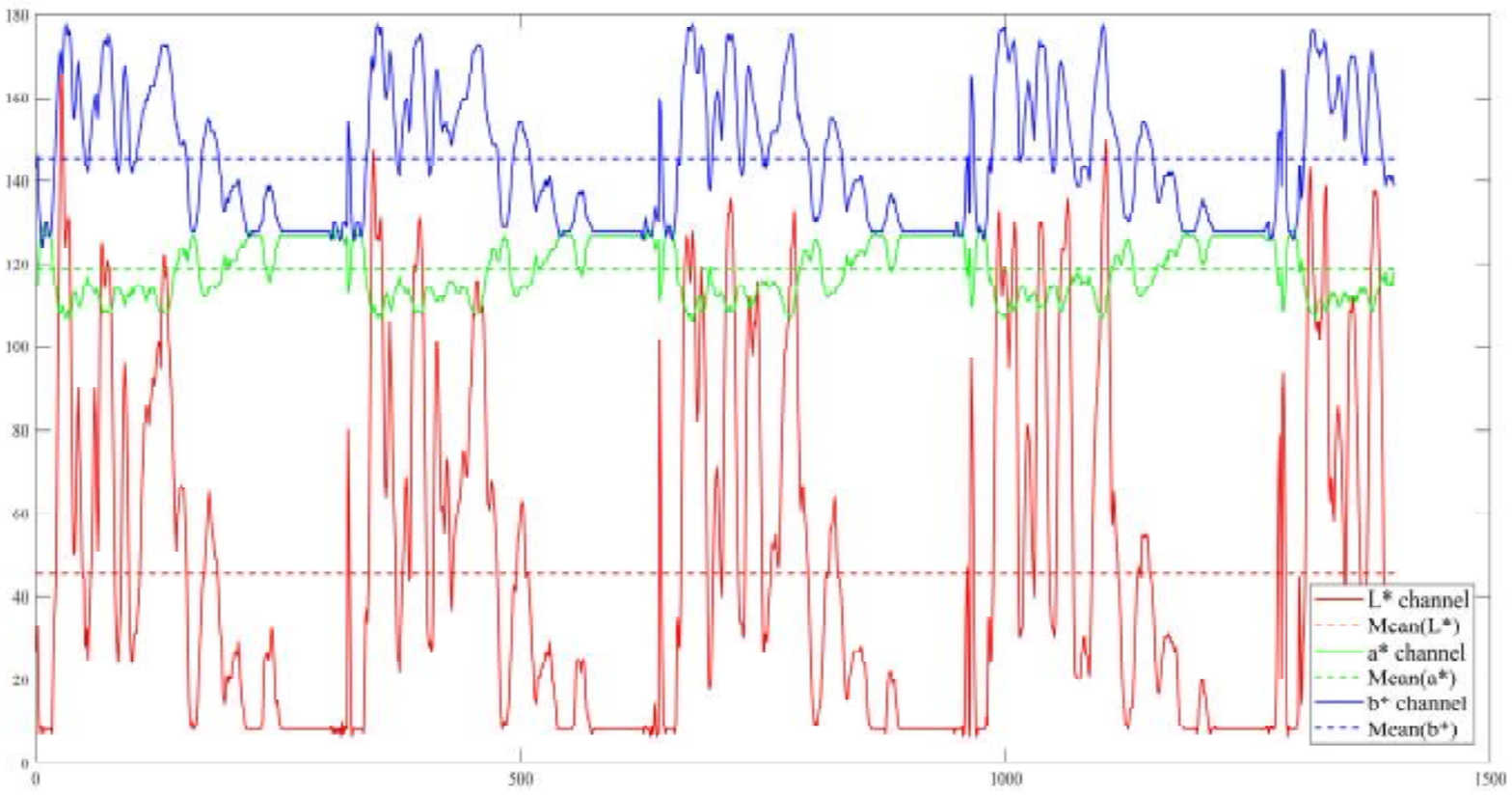} \label{f11a}
&
\includegraphics[width=7cm, height=6cm]{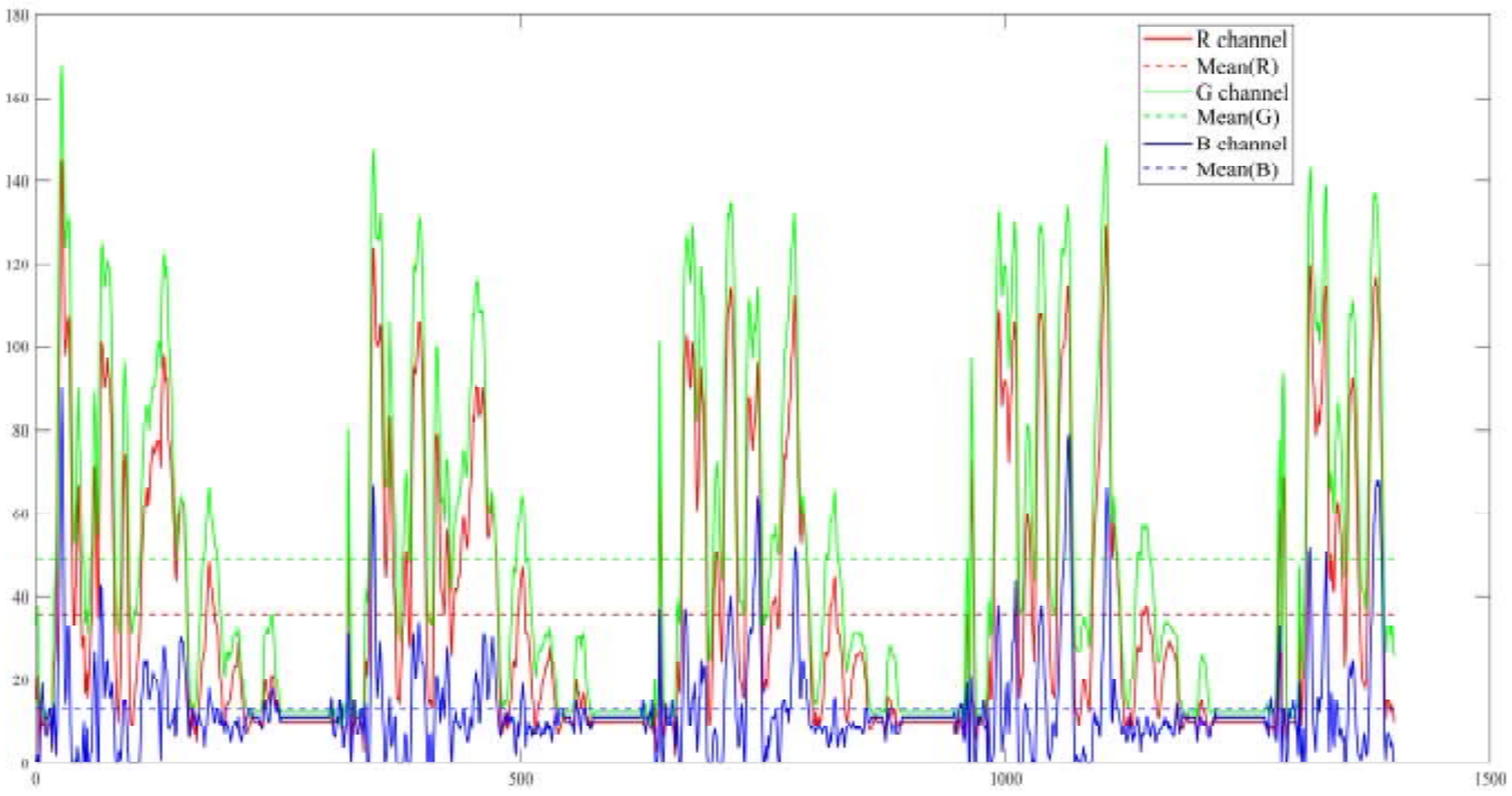} \label{f11b}
\\ (a) & (b)
\end{tabular}
\end{center}
\caption{The distributions of pixel values of the "Puppy" image in RGB and CIE L*a*b* color models.}
\label{f11}
\end{figure}

The distributions of pixel values of "Puppy" image in RGB and CIE L*a*b* color models are shown in Fig ~\ref{f11} (a) and (b). It can be observed that the pixel values in L*, a* and b* channels are generally larger than those in R, G and B, channels. And there is a huge difference of pixel values in each channel. In addition, the average value of pixels in each channel can be used to illustrated that the color gamut of L*a*b* is wider than the color gamut of RGB. And the larger difference of pixel values is very useful to extract edge information. Therefore, the necessity of using L*a*b* color space and its advantages of color representation are proved.

\subsection{The multi-scale and multi-directional Gabor filter}
\label{sec:3}
\begin{figure}
\begin{center}
\begin{tabular}{cc}
\includegraphics[width=7cm, height=6cm]{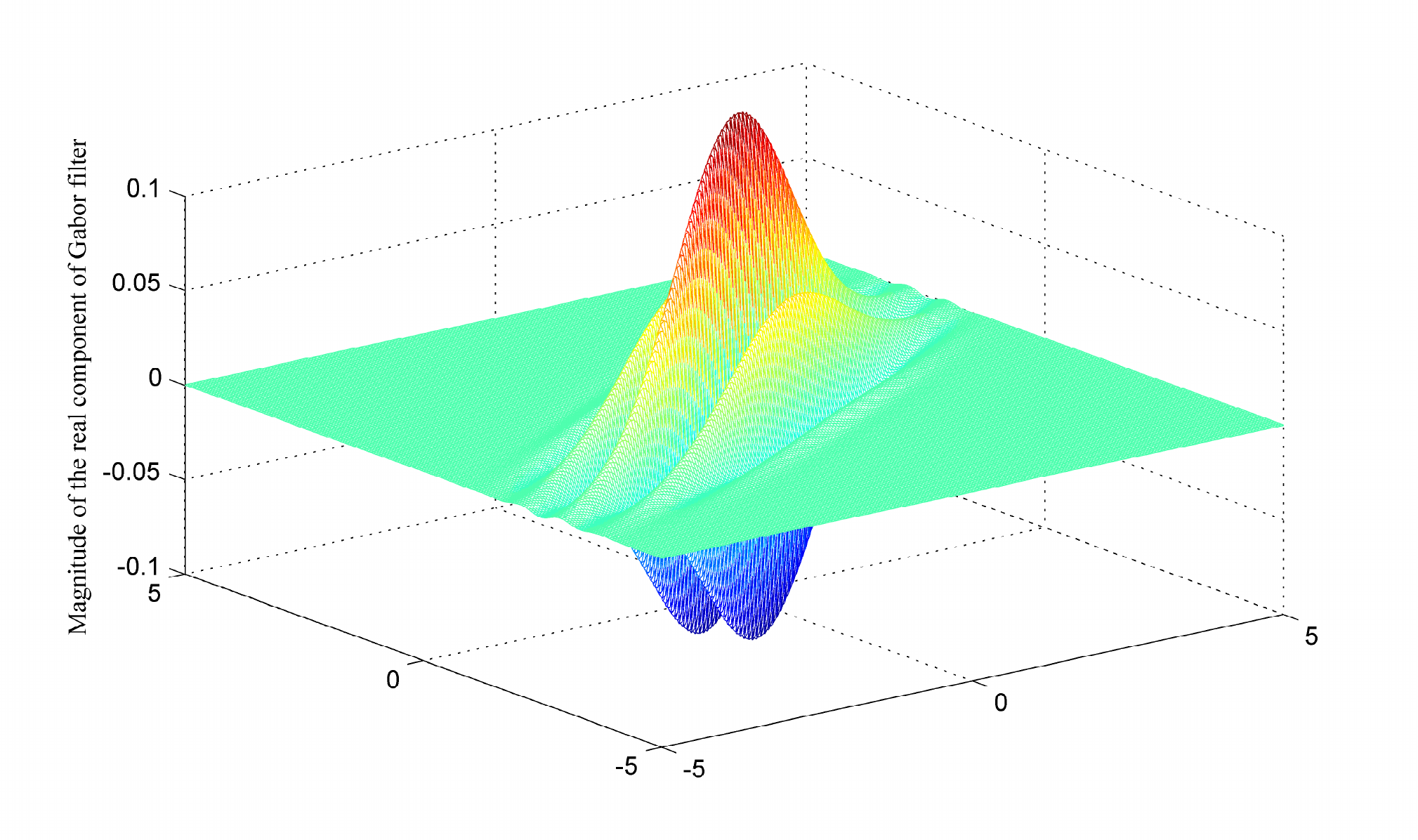}&
\includegraphics[width=7cm, height=6cm]{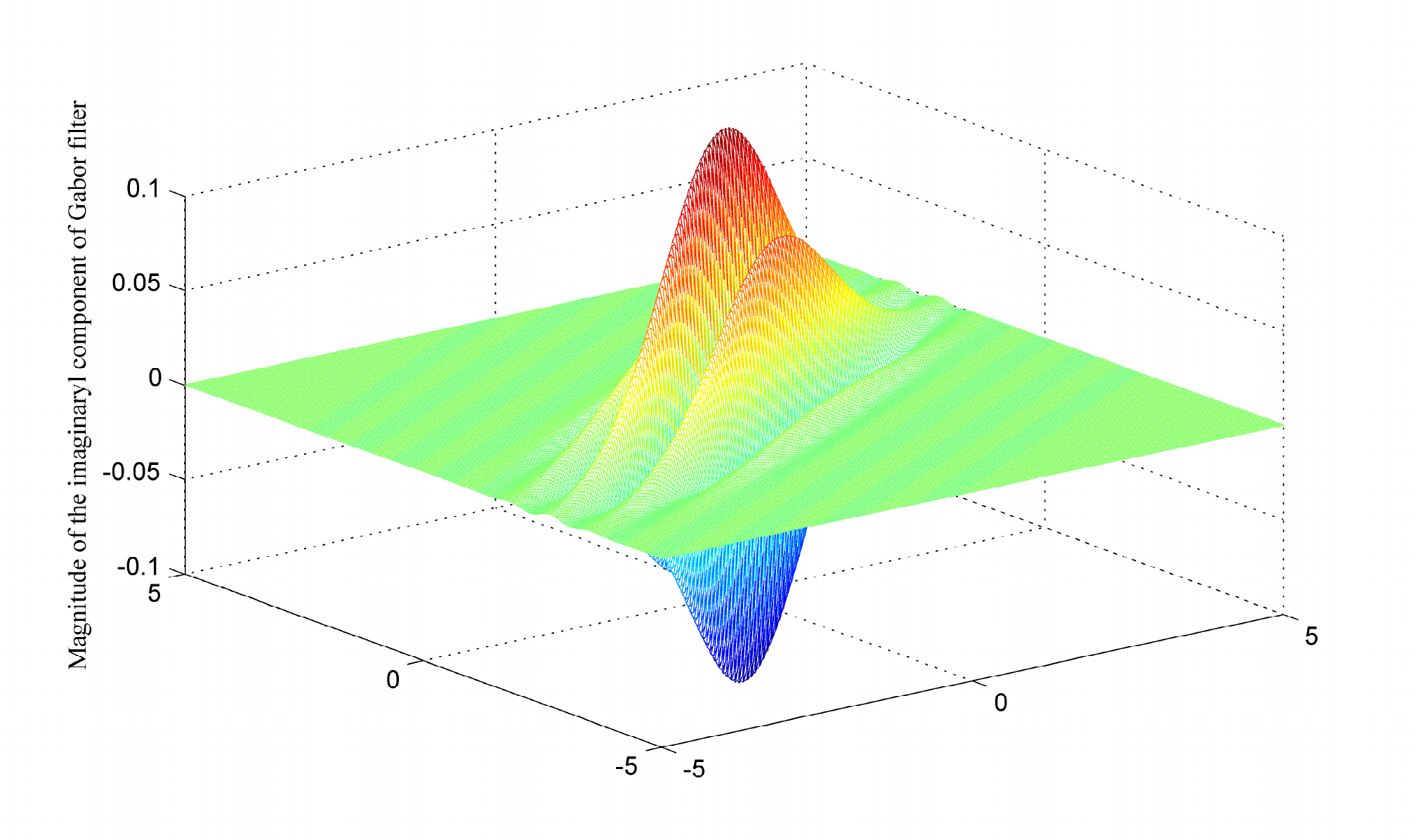}
\\
(a) & (b)
\end{tabular}
\end{center}
\caption{Examples of real and imaginary components of the Gabor filter. (a) A real component of Gabor filter with the $\gamma=1.3$, $\eta=2.5$, $f=0.3$ and $\theta=\frac{2}{3}\pi$; (b) An imaginary component of Gabor filter with the $\gamma=1.3$, $\eta=2.5$, $f=0.3$ and $\theta=\frac{2}{3}\pi$.}
\label{f2}
\end{figure}

The Gabor filter, as a liner filter, is similar to perception of the human visual system because of the frequency and orientation representations. Therefore, it is applied to the edge detection of the image. The impulse response of Gabor filter can be regarded as a sine/cosine wave multiplied by a Gaussian function. The complex, real and imaginary expressions of the Gabor filter are respectively

\begin{equation*}
\begin{split}
g(x, y; f, \theta)= \frac{f^2}{\pi\gamma\eta}\exp\left(-\left(\frac{f^2}{\gamma^2}x'^2+\frac{f^2}{\eta^2}y'^2\right)\right)\exp(j2\pi fx'),
\end{split}
\end{equation*}

\begin{equation*}
\begin{split}
g_{r}(x, y; f, \theta)= \frac{f^2}{\pi\gamma\eta}\exp\left(-\left(\frac{f^2}{\gamma^2}x'^2+\frac{f^2}{\eta^2}y'^2\right)\right)\cos(2\pi fx'),
\end{split}
\end{equation*}

\begin{equation*}
\begin{split}
g_{i}(x, y; f, \theta)= \frac{f^2}{\pi\gamma\eta}\exp\left(-\left(\frac{f^2}{\gamma^2}x'^2+\frac{f^2}{\eta^2}y'^2\right)\right)\sin(2\pi fx'),
\end{split}
\end{equation*}

\begin{equation*}
x'=x\cos\theta+y\sin\theta,
\end{equation*}

\begin{equation}
y'=-x\sin\theta+y\cos\theta,
\label{eq4}
\end{equation}
where $\gamma$ and $\eta$ are constants, which represent the sharpness along the horizontal and vertical axis, respectively. The central frequency of the filter is $f$. And the $\theta$ is the rotation angle.
It can be observed from Equation~(\ref{eq4}) that the scale of the filter is related to the center frequency $f$. A low center frequency indicates a large-scale Gabor filter, which has a good noise robustness. While a high center frequency indicates a small-scale Gabor filter, which has better localization of the edges in the image. The real and imaginary components of Gabor filter is shown in Fig ~\ref{f2} (a) and (b) respectively.

\begin{figure}
\begin{center}
\begin{tabular}{ccc}
\includegraphics[width=4cm, height=4cm]{231.pdf} \label{22a} &
\includegraphics[width=4cm, height=4cm]{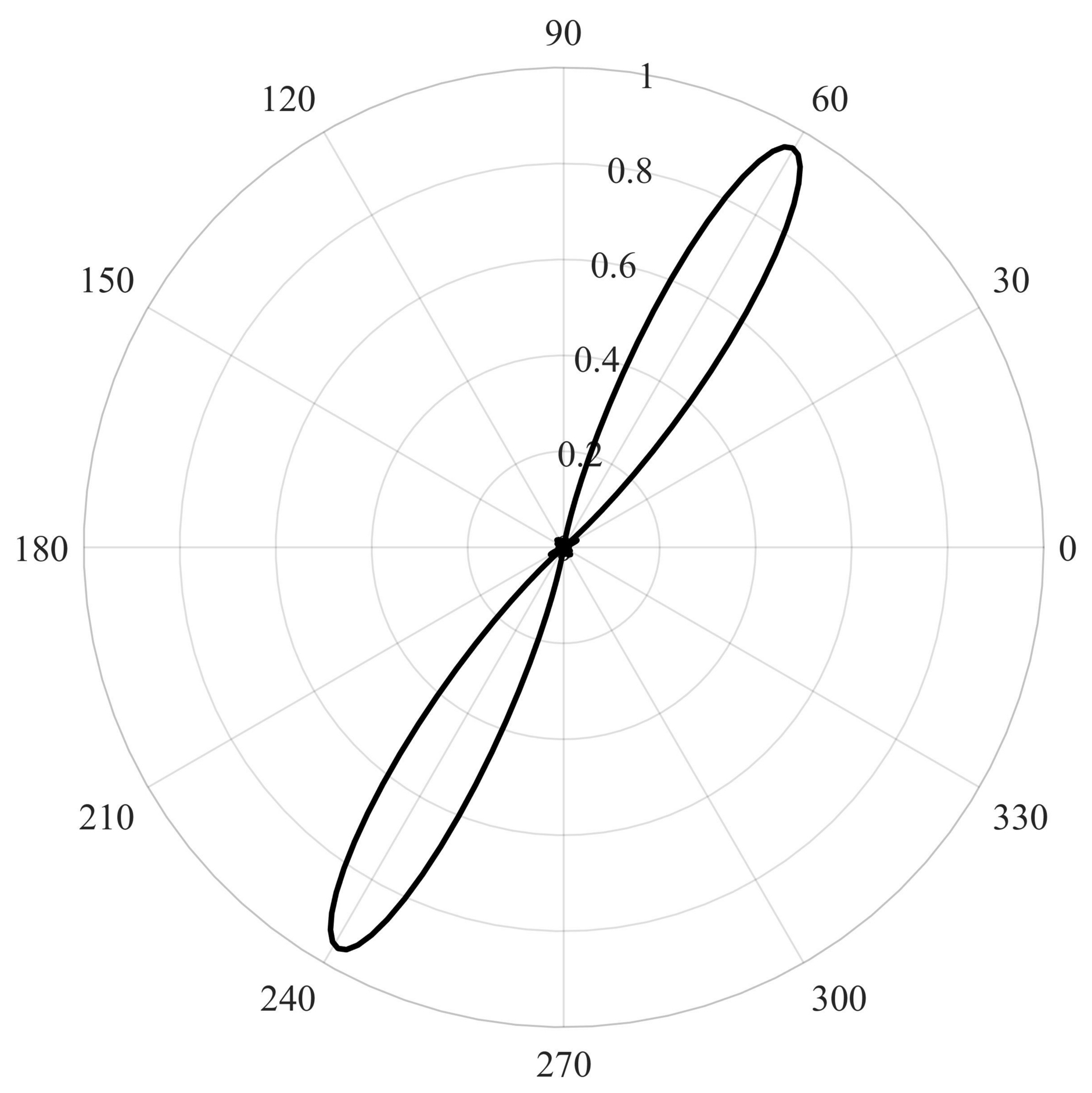} \label{22b} &
\includegraphics[width=6cm, height=4cm]{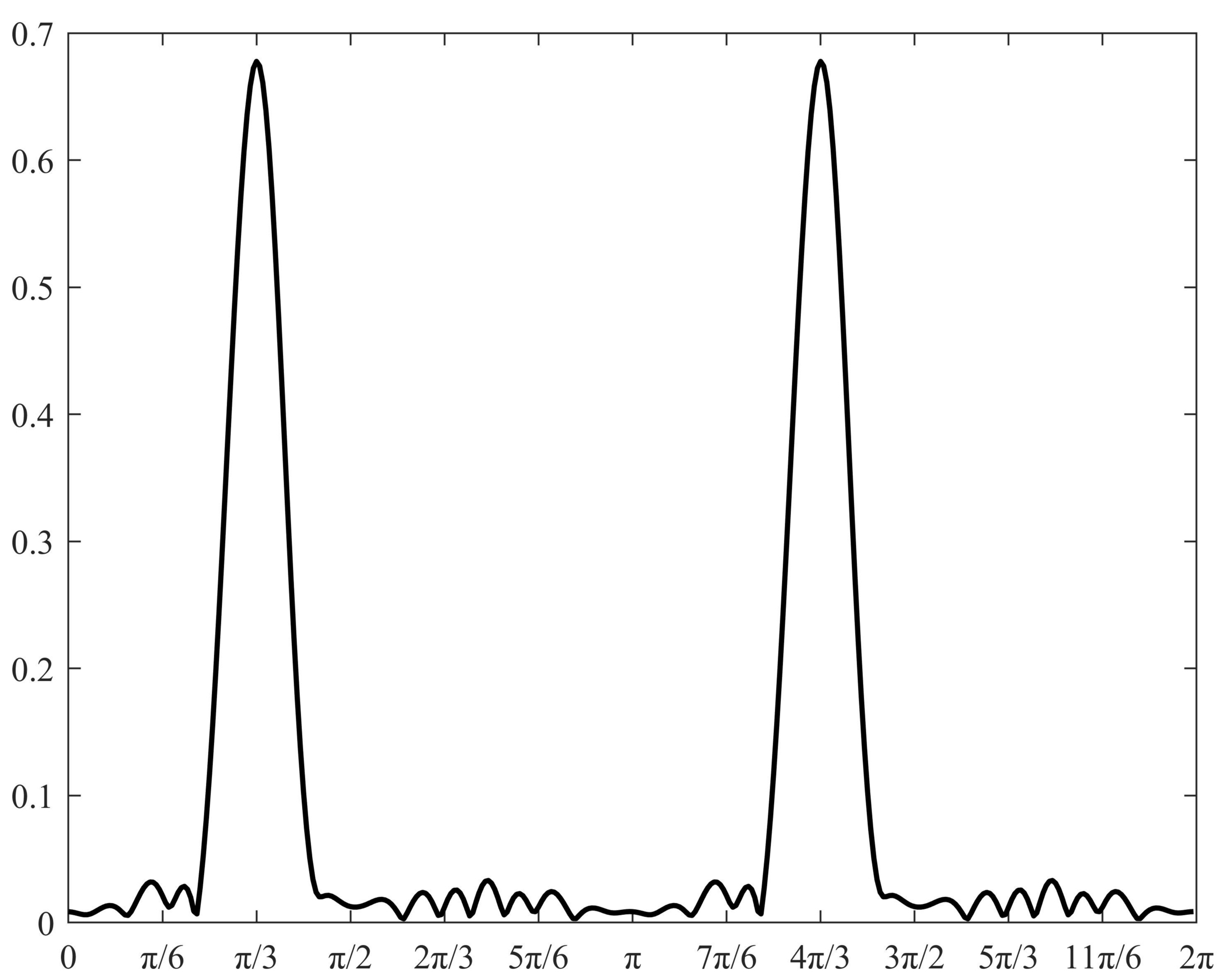} \label{22c}
\\ (a) & (b) & (c) \\

\includegraphics[width=4cm, height=4cm]{234.pdf} \label{22d} &
\includegraphics[width=4cm, height=4cm]{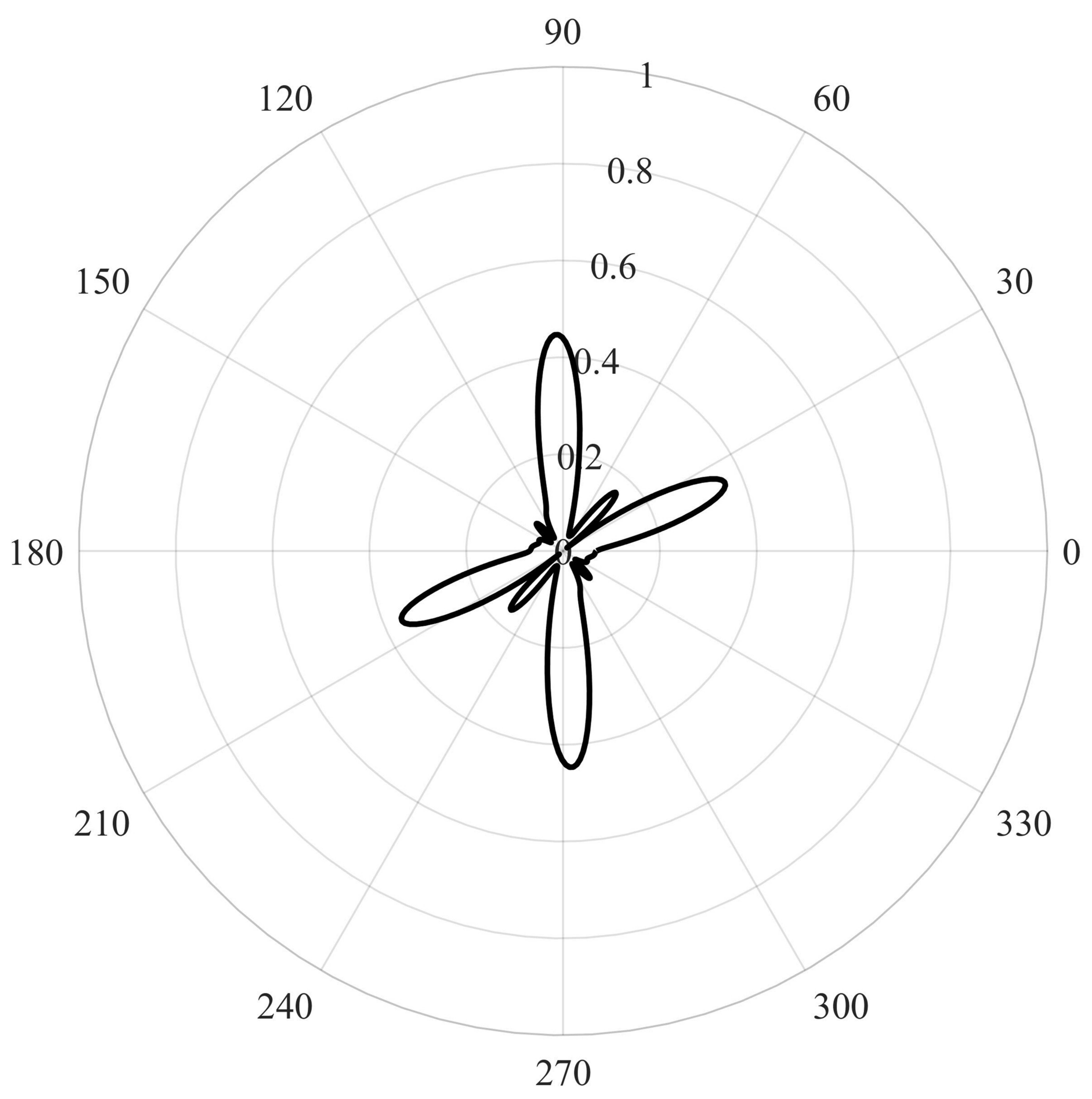} \label{22e} &
\includegraphics[width=6cm, height=4cm]{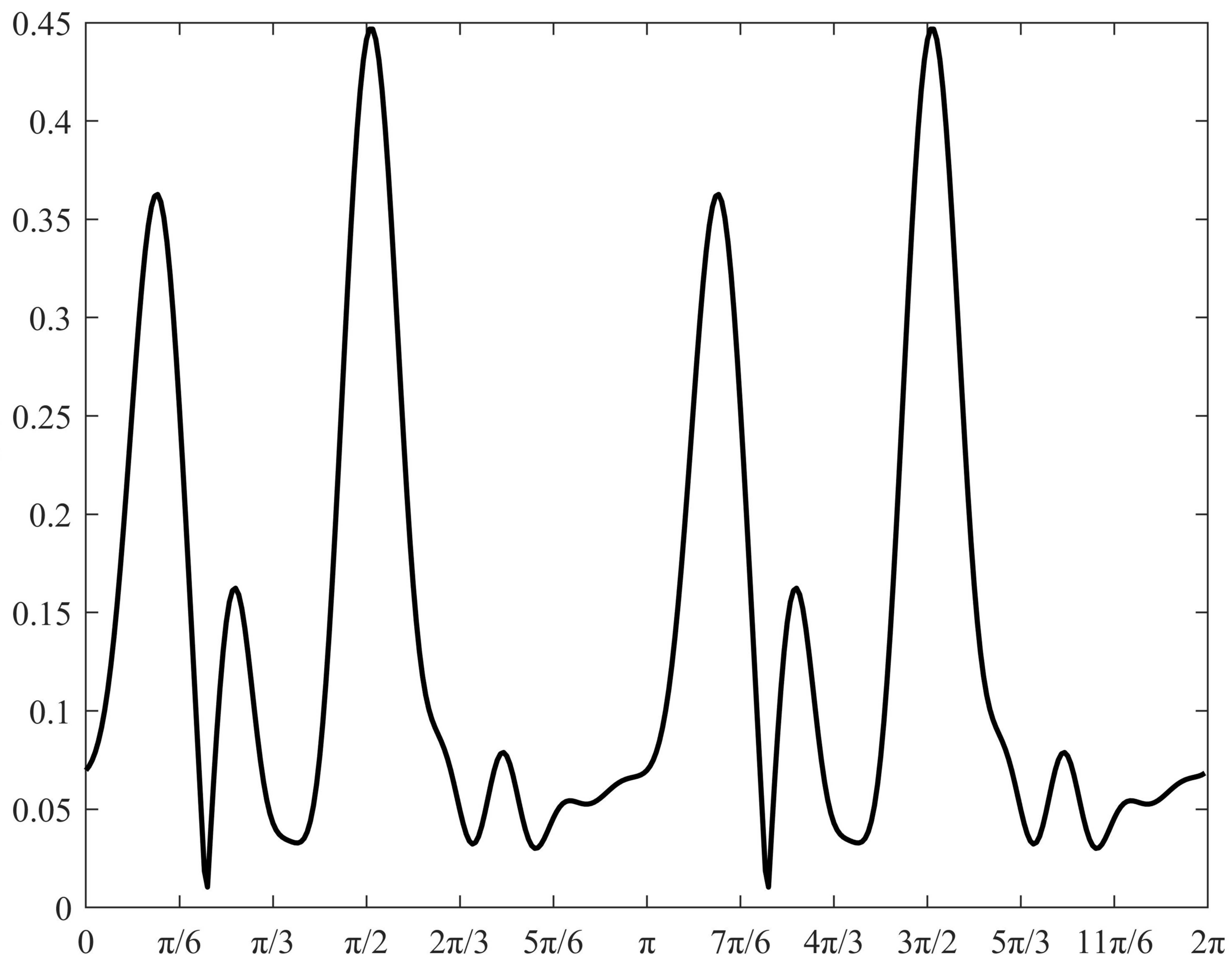} \label{22f}
\\ (d) & (e) & (f) \\

\includegraphics[width=4cm, height=4cm]{237.pdf} \label{22g} &
\includegraphics[width=4cm, height=4cm]{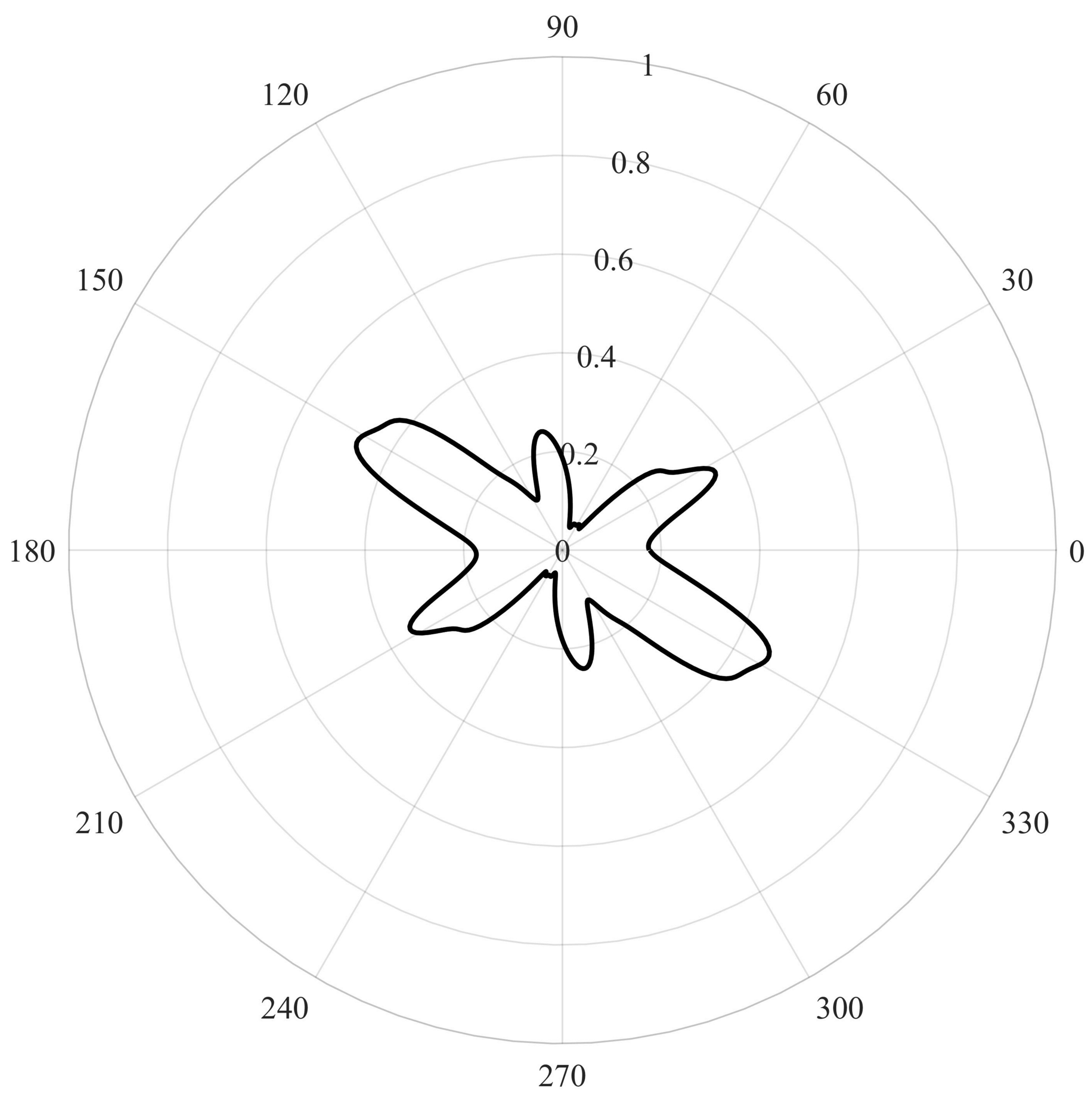} \label{22h} &
\includegraphics[width=6cm, height=4cm]{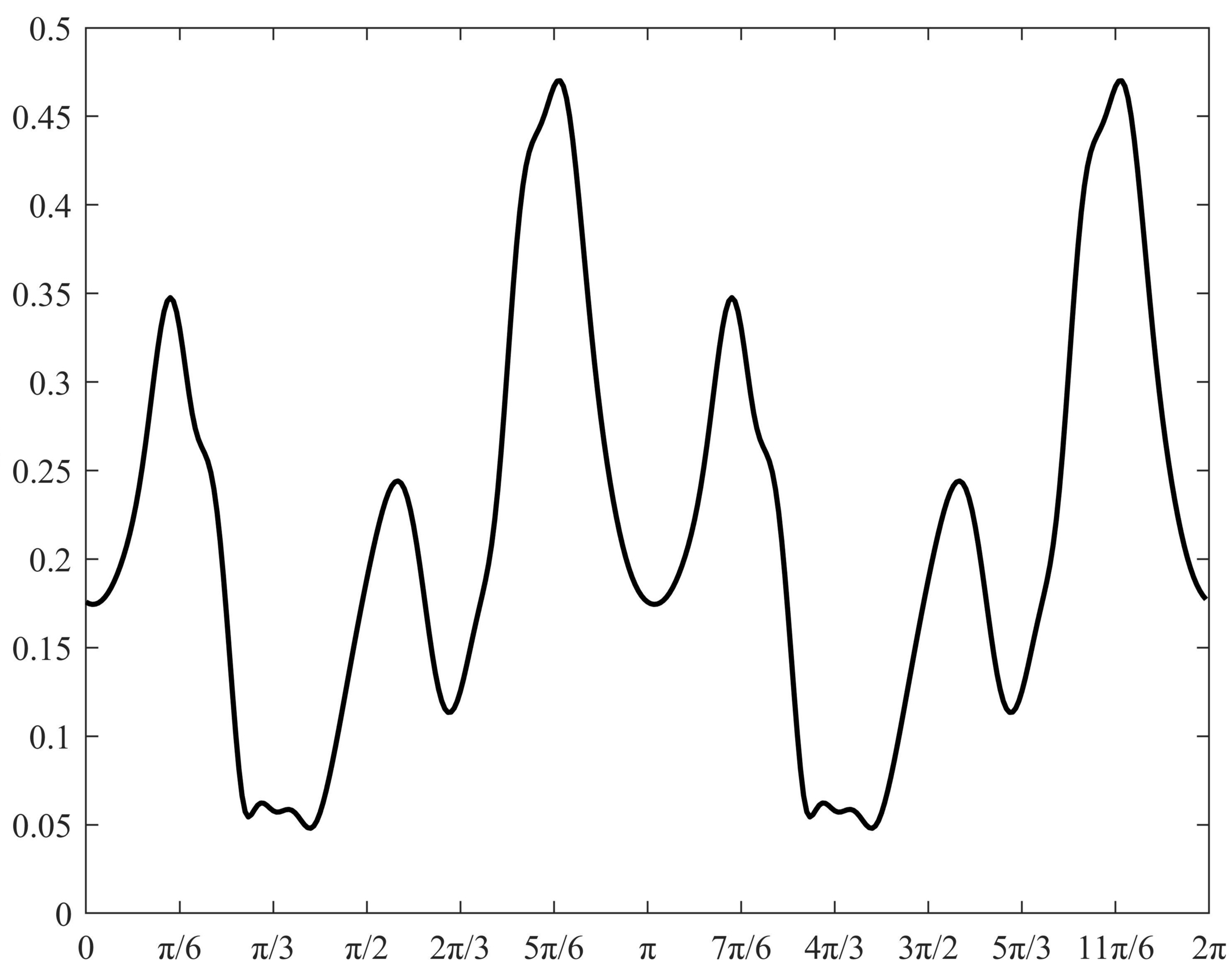} \label{22i}
\\ (g) & (h) & (i) \\

\includegraphics[width=4cm, height=4cm]{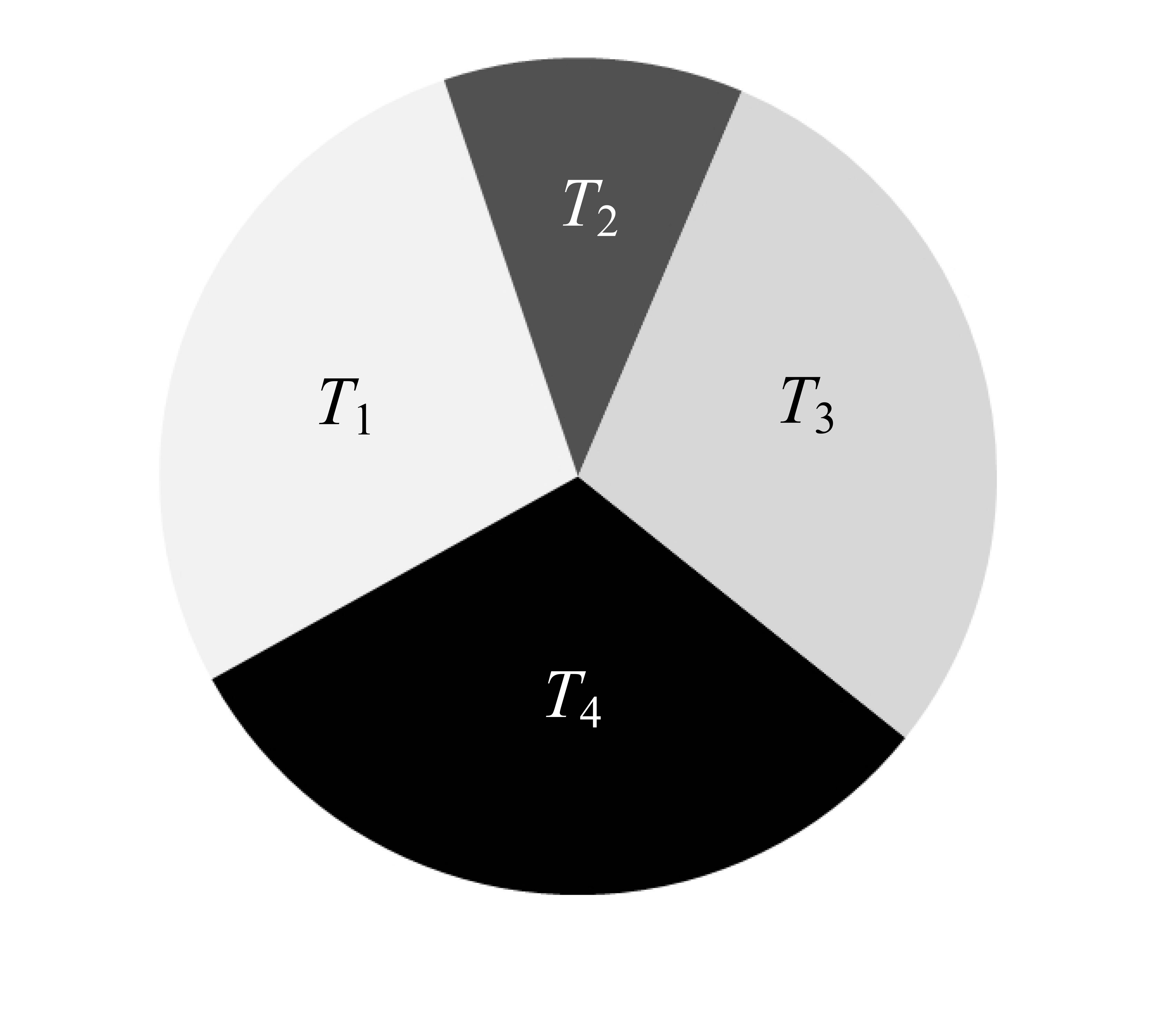} \label{22j} &
\includegraphics[width=4cm, height=4cm]{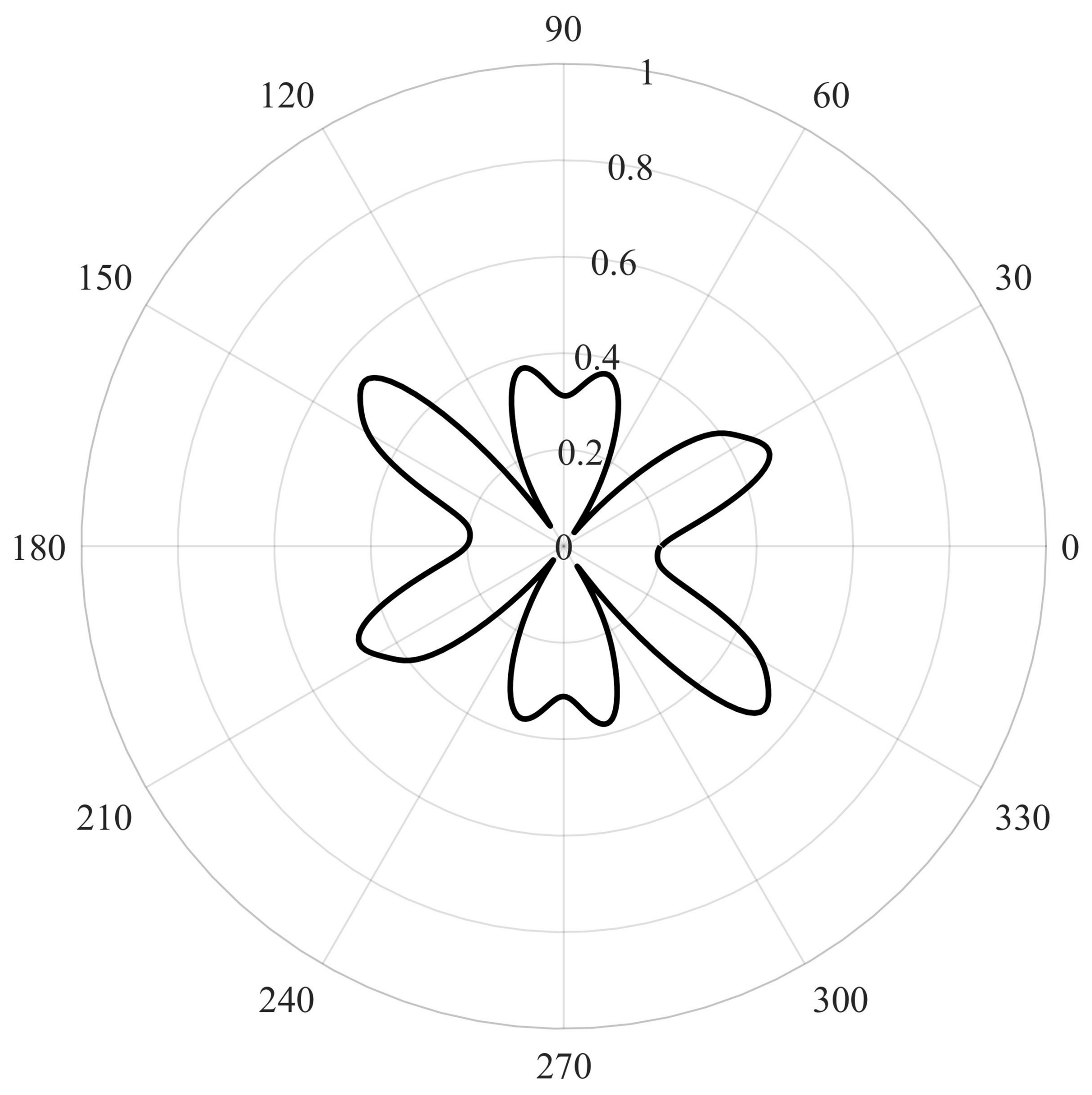} \label{22k} &
\includegraphics[width=6cm, height=4cm]{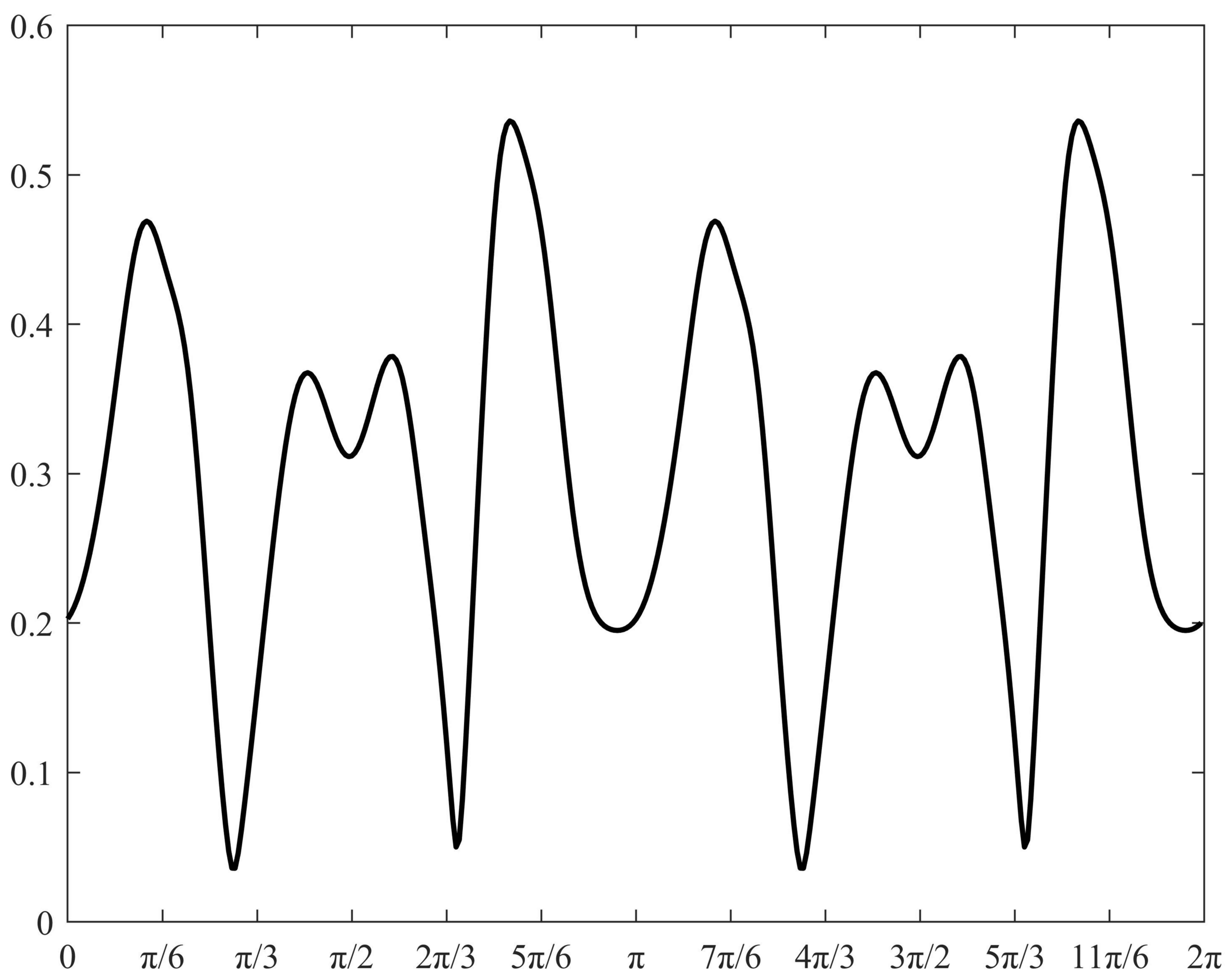} \label{22l}
\\ (j) & (k) & (l) \\

\includegraphics[width=4cm, height=4cm]{2313.pdf} \label{22m} &
\includegraphics[width=4cm, height=4cm]{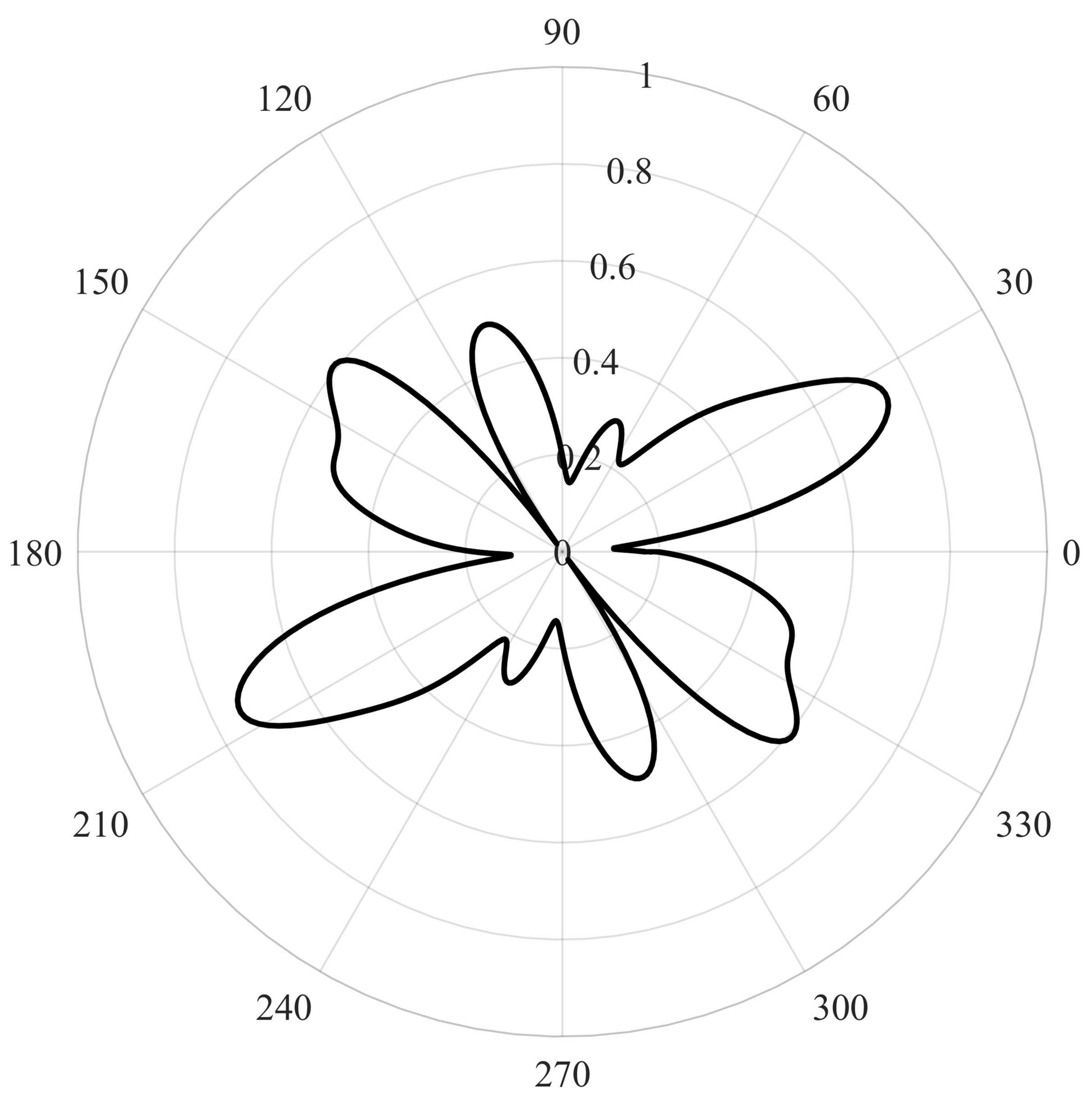} \label{22n} &
\includegraphics[width=6cm, height=4cm]{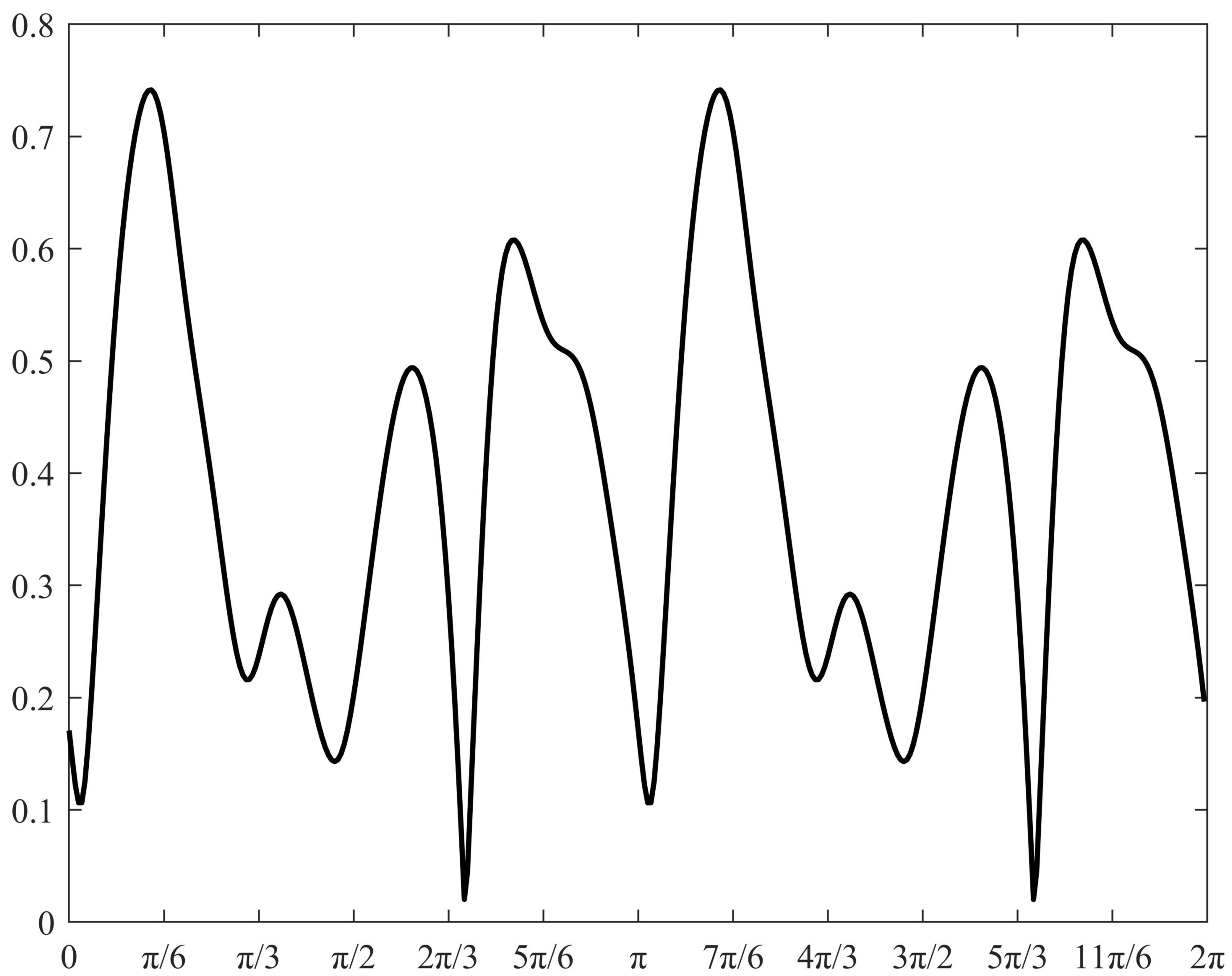} \label{22o}
\\ (m) & (n) & (o)

\end{tabular}
\end{center}
\caption{Magnitude responses of different types of edges and Gabor filters.}
\label{f22}
\end{figure}

From left to right, a step edge, a simple angular edge, a Y-shaped edge, a X-shaped edge and a star-shaped edge are shown in the first column of the Fig ~\ref{f22} (a), (d), (g), (j), (m). And the magnitude responses of different types of edges and Gabor filters in polar coordinates and cartesian coordinates are shown in the second and third columns of the Fig ~\ref{f22}. The Gabor filter is anisotropic, which can effectively extract the information of edge. And the relatively different responses can be observed in the area $T_{i}$ with different gray values. The multi-scale and multi-directional Gabor filter is extremely powerful for the task of edge detection.

From left to right, a step edge, a simple angular edge, a Y-shaped edge, a X-shaped edge and a star-shaped edge are shown in the first column of the Fig ~\ref{f22} (a), (d), (g), (j), (m). And the magnitude responses of different types of edges and Gabor filters in polar coordinates and cartesian coordinates are shown in the second and third columns of the Fig ~\ref{f22}. The Gabor filter is anisotropic, which can effectively extract the information of edge. And the relatively different responses can be observed in the area $T_{i}$ with different gray values. The multi-scale and multi-directional Gabor filter is extremely powerful for the task of edge detection.

Considering the superiority of the multi-scale filter, the scale factor $\sigma$ is added in the Gabor filter. For an input image $I (\mathbf{x})$, $\mathbf{x}=(x,y)^T$, the local features of the image at different scales are attained by using the multi-scale Gabor filter $g(x, y; f, \theta, \sigma)$to smooth the image:

\begin{equation}
\begin{aligned}
I_{\sigma}(\mathbf{x})&=I\otimes g(x, y; f, \theta, \sigma)\\
&=\iint^{+\infty}_{-\infty}I(x_{\tau},y_{\tau})\cdot g(x-x_{\tau}, y-y_{\tau}; f, \theta, \sigma)dx_{\tau}dy_{\tau} ,
\end{aligned}
\label{eq5}
\end{equation}
where superscript $T$ represents matrix transpose and symbol ''$\otimes$''  represents a convolution operation.

\begin{figure}
\begin{center}
\begin{tabular}{c}
\includegraphics[width=15cm, height=4.5cm]{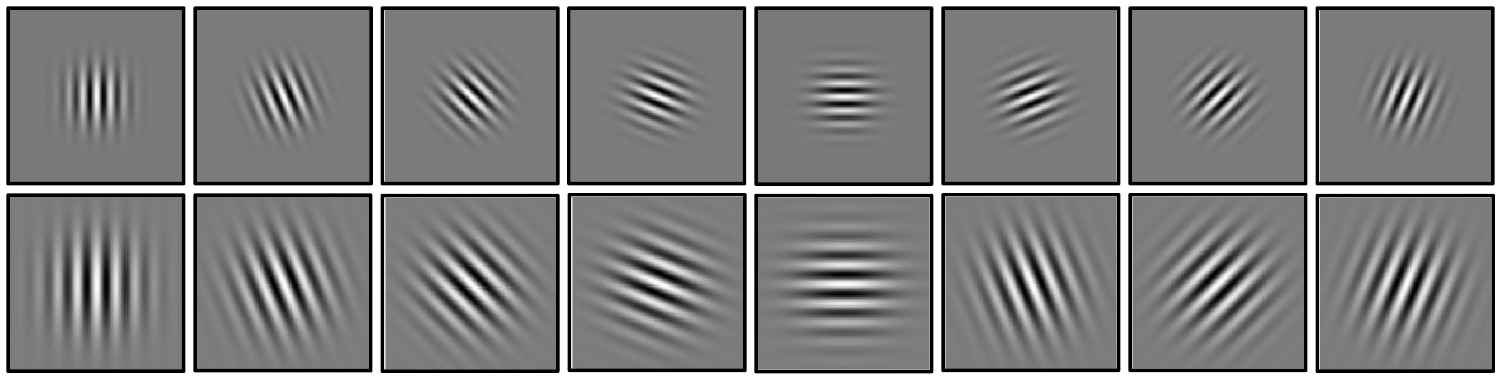}
\end{tabular}
\end{center}
\caption{A set of Gabor filters with eight orientations(From left to right, the orientations are $0$, $\frac{\pi}{8}$, $\frac{\pi}{4}$, $\frac{3\pi}{8}$, $\frac{\pi}{2}$, $\frac{5\pi}{8}$, $\frac{3\pi}{4}$, $\frac{7\pi}{8}$) in the first column to the eighth column, respectively. The central frequency of the first row is $f_1$ ($f_1=0.1$) and the second row is $f_2$ ($f_2=0.2$). $\gamma=1$,$\eta=2$.}
\label{f3}
\end{figure}

The discrete operation is required by the continuous Gabor filter in Equation~(\ref{eq5}) when the input images are 2D discrete signals in the integer lattice $Z^2$. In addition, the filter along different orientations need to be used to extract local variation information around a pixel. A set of discretized multi-scale Gabor filters can be obtained by taking different center frequencies $f$:

\begin{gather}
\begin{aligned}
\varphi(u,v;f,\sigma,k)=&\frac{f_{\sigma}^2}{\pi\gamma\eta}\exp\left(-\left(\frac{f_{\sigma}^2}{\gamma^2}u'^2+\frac{f_{\sigma}^2}{\eta^2}v'^2\right)\right)\\&\cdot\exp(j2\pi f_{\sigma}u')
\end{aligned} \notag \\
u'=u\cos{\theta_{k}}+v\sin{\theta_{k}}, \notag \\
v'=-u\sin{\theta_{k}}+v\cos{\theta_{k}}, \notag \\
\theta_{k}=\frac{\pi k}{K}, k=0, 1, ..., K-1
\label{eq6}
\end{gather}

where $K$ is the number of directions, $\theta_{k}$ is the k-th orientation, and $f_{\sigma}$ is the center frequency of the s-th scale. The Gabor filters with eight orientations at two scales are shown in Fig ~\ref{f3}.

In view of the aforementioned formula derivation, for an input discrete image $I(\mathbf{z})$, $\mathbf{z}=(u,v)^{T}$, the magnitude response of the discrete Gabor filters along the orientation $\theta_{k}$ at the scale $f_{\sigma}$ is calculated by the convolution operator

\begin{equation}
\begin{aligned}
\psi(u,v;f,\sigma,k)&=I\otimes\varphi(u,v;f,\sigma,k)\\
&=\sum_{u_{p}}\sum_{v_{q}}I(u-u_{p},v-v_{q})\varphi(u,v;f,\sigma,k).
\end{aligned}
\label{eq7}
\end{equation}

As a result, a set of the Gabor filters along the different orientations at the different scales can reflect the intensity variation around the edge pixels completely.

\section{A new color edge detection measure}

\label{sec:4}
In this section, using the multi-scale Gabor filter, the ESMs of the three channels (L*, a* and b*) are presented. Furthermore, the fused ESM is obtained. Finally, a new color edge detection measure is derived by embedding the fused ESM in the route of the Canny detector which has good edge detection accuracy and noise robustness.

\subsection{The ESMs of the color image and the proposal of fused ESM}
\label{sec:5}
Canny~\cite{canny1986a} pointed out that the optimal filter should have three criteria: (\romannumeral1) Good detection, There should be a low probability of failing to mark real edge points, and low probability of falsely marking non-edge points; (\romannumeral2) Good localization. The points marked as edge points
by the operator should be as close as possible to the center of the true edge. (\romannumeral3) Only one response to a single edge. This is implicitly captured in the first criterion since when there are two
responses to the same edge, one of them must be considered false. Meanwhile a set of optimal filter banks was obtained through demonstration. However, the Canny detector also has some shortcomings. Smoothing the edges with a large-scale Gaussian kernel will cause the edge to blur and lose a lot of details. This situation is especially obvious in the texture area, while the small-scale Gaussian kernel is very sensitive to noise. It is the inherent defect of Canny detection operator. Under this situation, the scale and the orientation should be considered for the optimal filter design, which is regard as a indispensable step in edge detection.

From the aforementioned analysis, the multi-scale Gabor filter is noise-robust and high edge accuracy in the case of the scale and orientation factors. The $max$ operator in Equation~(\ref{eq8}) performs a simple selective smoothing~\cite{alvarez1992image} of an image and the ''optimal'' smooth kernel to each pixel is selected from the K predefined in the Gabor filters. Image smoothing is indispensable for edge detection of noisy images. The ESM from the multi-scale Gabor filter of an image is defined by

\begin{equation}
\zeta(\mathbf{z})=\max_{k=0,1,2,\ldots,K-1}\left\{|\psi(u,v;f,\sigma,k)|\right\},
\label{eq8}
\end{equation}

When an color image is converted from the RGB space to the L*a*b space, the L*, a* and b* color characteristics can be acquired. Each color characteristic is expressed as a 2-D matrix. The use of Equation~(\ref{eq8}) smoothes the L*, a* and b* channels, respectively. Then, according the ESMs of the three channels, an overlay operation is adopted, which can get better filtering effects. In the end, the fused ESM can be calculated by

\begin{equation}
\xi(\mathbf{z})=\sqrt[n]{\zeta_{1}(\mathbf{z})\cdot\zeta_{2}(\mathbf{z})...\cdot\zeta_{n}(\mathbf{z})}.
\label{eq9}
\end{equation}

\begin{figure}
\begin{center}
\begin{tabular}{c}
\includegraphics[width=15cm, height=7cm]{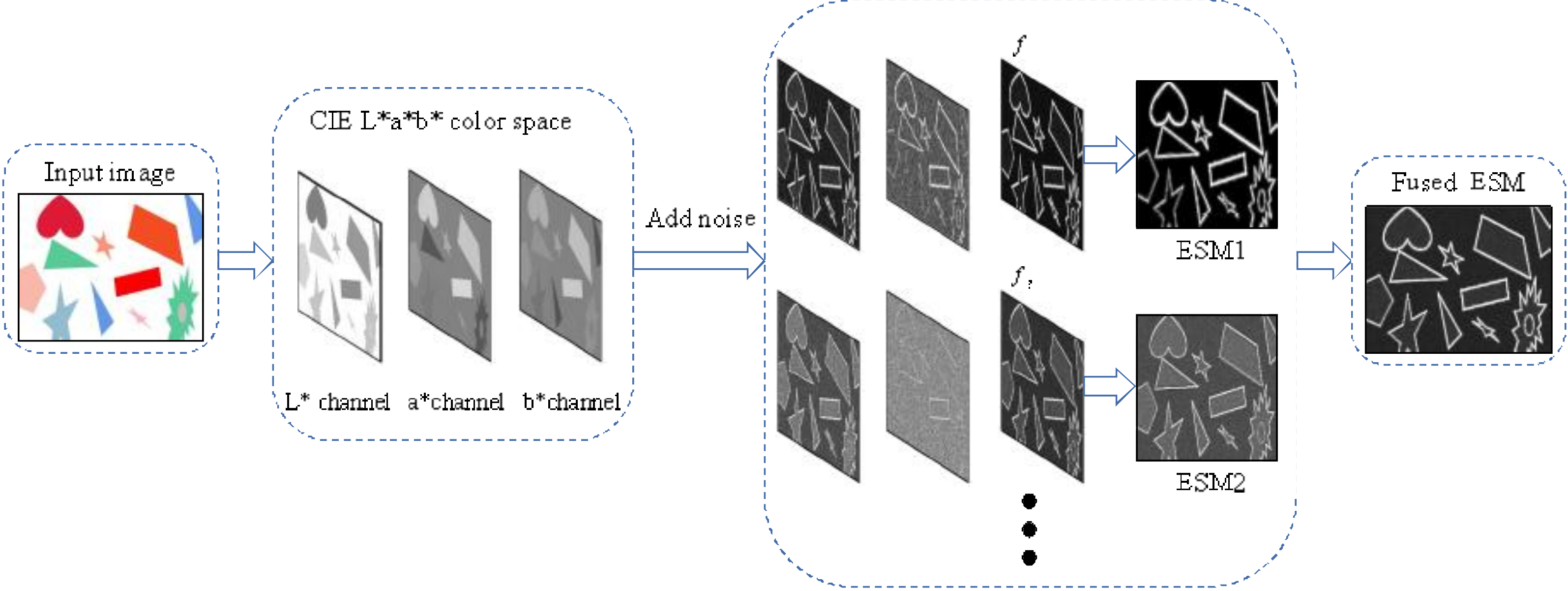}
\end{tabular}
\end{center}
\caption{An example of the flow chart of the fused ESM.}
\label{f4}
\end{figure}

The fused ESM based on Equation(\ref{eq9}) has the ability to solve the impact of the blurring effect caused by chromaticity and maintain high edge accuracy, and noise robustness. Under this approach, the proposed edge detection measure is defined as the fused ESM of the multi-scale Gabor filter.
An example of the flow chart of the fused ESM is shown in Fig ~\ref{f4}.

\begin{figure}
\begin{center}
\begin{tabular}{cc}
\includegraphics[width=6cm, height=4cm]{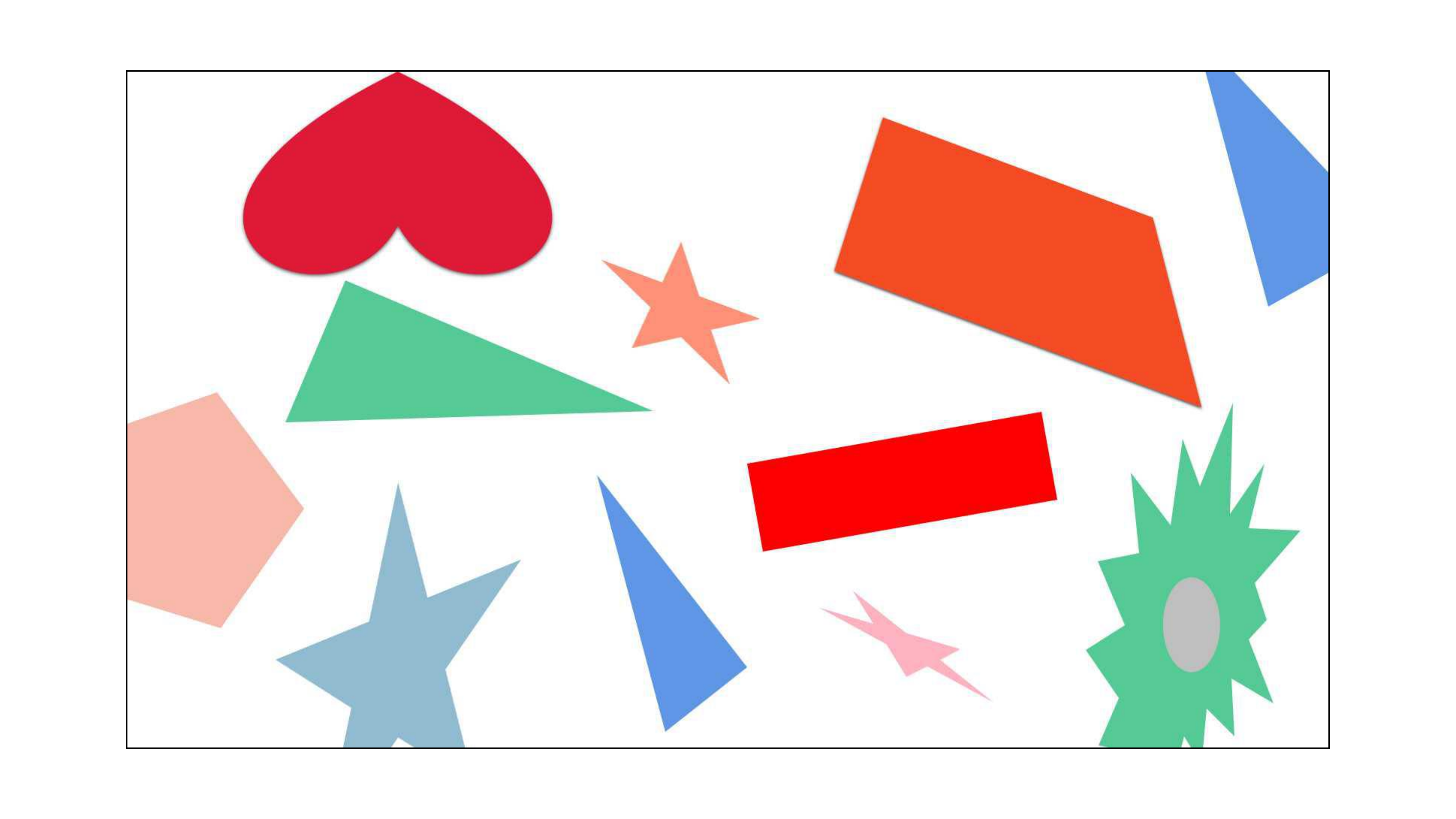}&
\includegraphics[width=6cm, height=4cm]{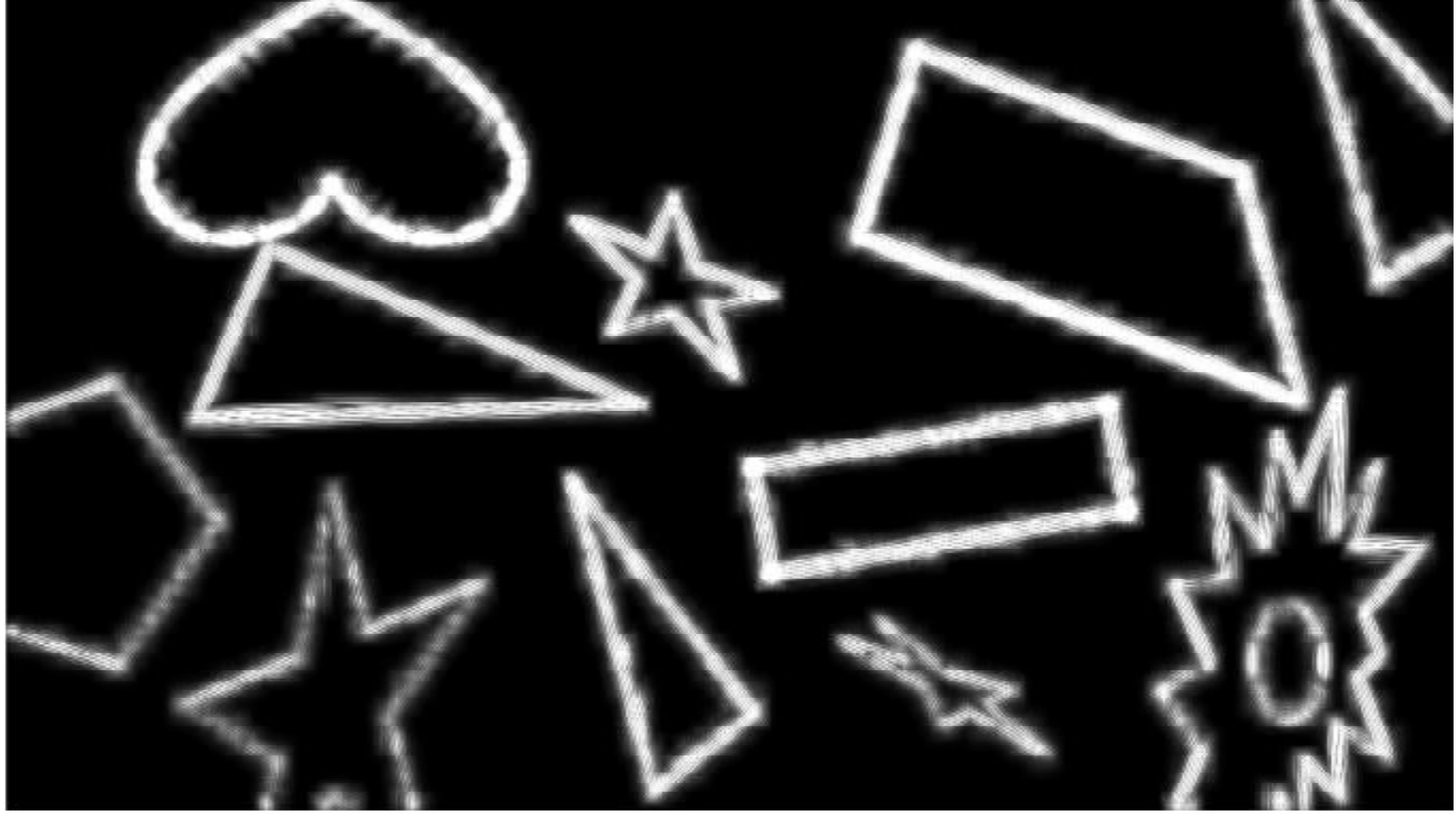}
\\(a) & (b)
\\
\includegraphics[width=6cm, height=4cm]{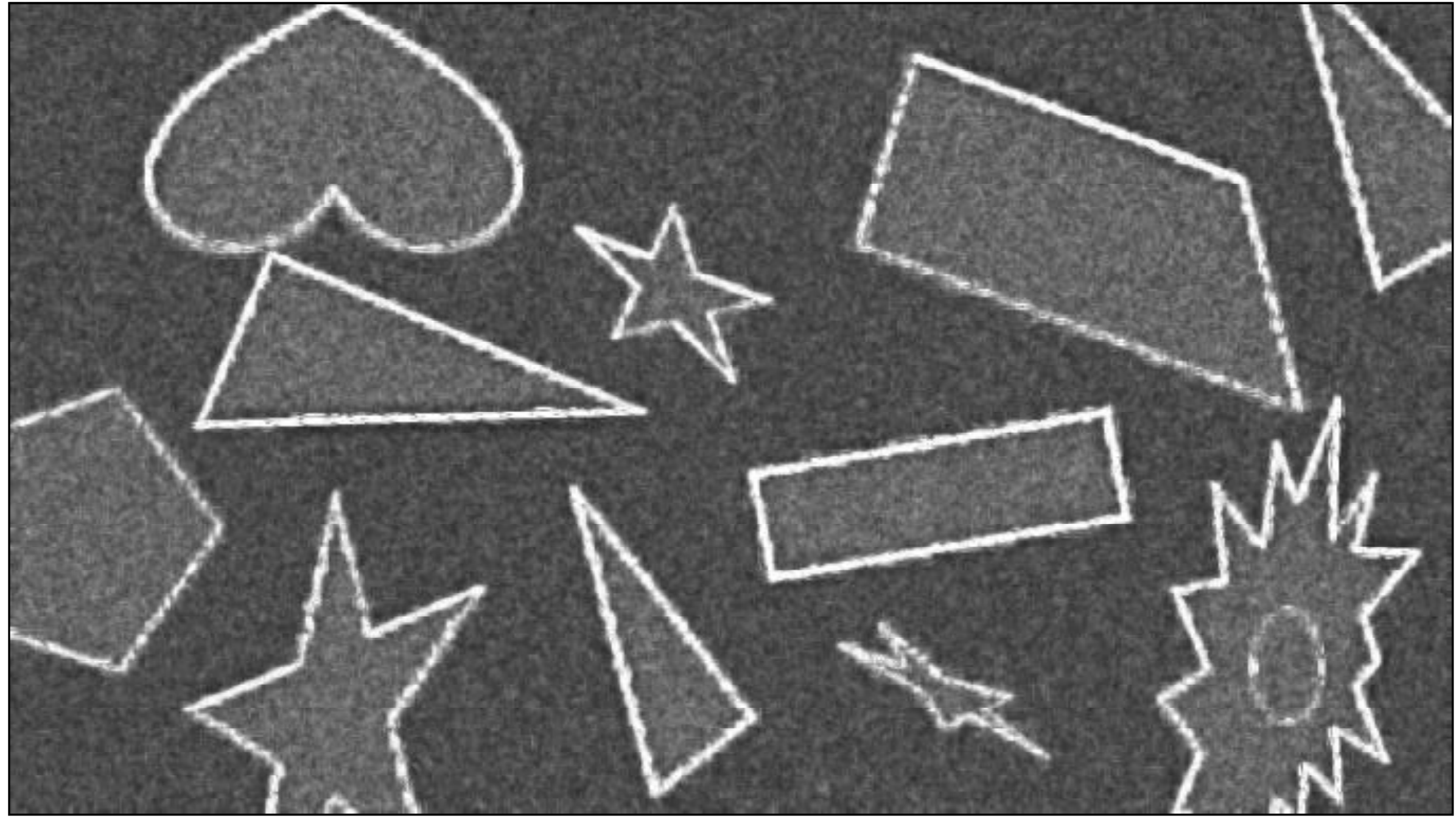}&
\includegraphics[width=6cm, height=4cm]{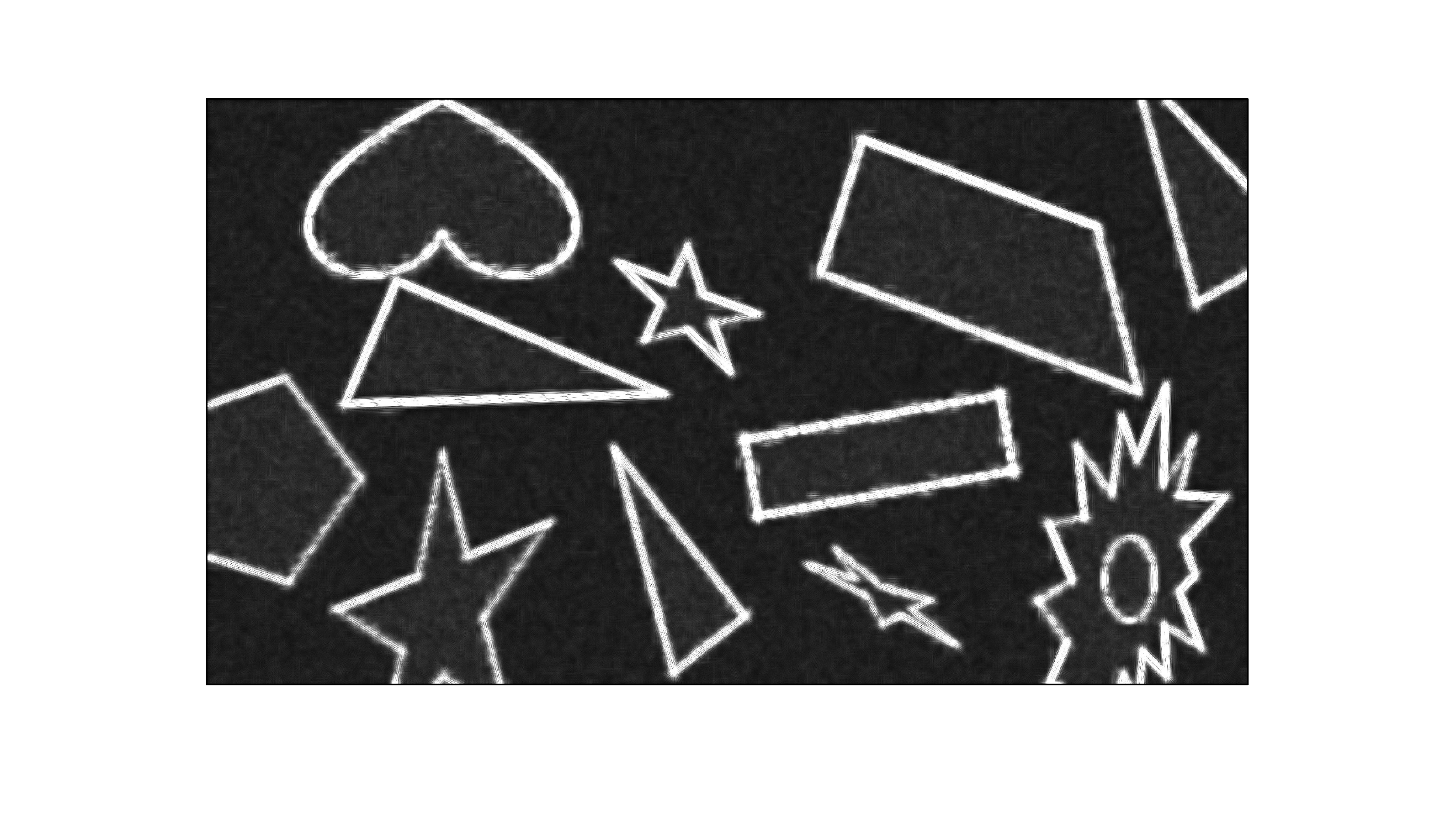}
\\(c) & (d)
\end{tabular}
\end{center}
\caption{Illustration of the ESMs of an image: (a) an original image; (b) ESM1 at low $f_{1}$ ($f_{1}=0.2$) is corrupted by Gaussian noise ($\varepsilon_{w}=15$); (c) ESM2 at high $f_2$ ($f_{2}=2$) is corrupted by Gaussian noise ($\varepsilon_{w}=15$); (d) fused ESM.}
\label{f5},
\end{figure}

Comparing the ESM1, ESM2 and the fused ESM in Fig ~\ref{f5}, the merits of fused ESM are attained. The original image is shown in Fig.~\ref{f5} (a), which added noise with Gaussian standard deviation $\varepsilon_{w}=15$ before smoothed by multi-scale Gabor filters. The ESM1 with a low central
frequency $f_{1}$ ($f_{1}= 0.2$) is shown in Fig ~\ref{f5} (b). It can be observed from Fig ~\ref{f5} (b) that it is at a large scale so that there are more edge stretching on the ESM2 but it is rarely polluted by noise. As the scale decreases, the central frequency increases to $f_{2}$ ($f_{2}= 2$). It can be observed from Fig ~\ref{f5} (c) that the effect of noise on ESM is aggravated while the edge stretch effect of the ESM1 decreases and more details of pixels are obtained. The reason is that the ESM2 with a large scale is robust to noise. The fused ESM as shown in Fig ~\ref{f5} (d) inherits the merits of ESM1 and ESM2 at two different scales which has little edge stretch effect and it is noise robust.

\subsection{Proposed color edge detector using Gabor filter}
\label{sec:6}
The proposed edge detection algorithm first converts the color space from RGB to L*a*b*. Then, the multi-scale Gabor filters are used to smooth the images in each channels, and the magnitudes at every pixel along eight directions are obtained. Furthermore, $\zeta_{1}(\mathbf{z})$, $\zeta_{2}(\mathbf{z})$ ..., and $\zeta_{n}(\mathbf{z})$ are fused to be a new ESM. Finally the fused ESM is computed as the edge detection. The outline of the proposed algorithm is

(\romannumeral1) Convert color images from RGB space to L*a*b* space.

(\romannumeral2) Extract edge strength map from the each channel by multi-scale Gabor filters, and the fused ESM is attained by computation of ESMs in terms of Equation(\ref{eq9}).

(\romannumeral3) Calculate the average variation of the image $\overline{s}$ and local average variation $\overline{s}_{Local}(\mathbf{z})$.
The overall average change of an image can be calculated as
\begin{equation}
\overline{s}=\frac{1}{M\cdot N}\sum_{\mathbf{z}}\xi(\mathbf{z}),
\label{eq10}
\end{equation}
where $M\cdot N$ is the size of the input image, $\xi(\mathbf{z})$ is the value in Equation (\ref{eq9}). The local average variation of pixel $\mathbf{n}$ is defined as
\begin{equation}
\overline{s}_{Local}(\mathbf{z})=\frac{1}{W^2}\sum_{\tau\in Q}\eta(\mathbf{z}+\tau)),
\label{eq11}
\end{equation}
where $Q$ is a $W\times W$ squared window and $\tau$ is the changeable distance in the window. Simulating the visual system, the contrast equalization is used for the modified fused ESM, which is defined as
\begin{equation}
\tilde{\xi}(\mathbf{z})=\frac{\xi(\mathbf{z})}{\overline{s}+0.5\overline{s}_{Local}(\mathbf{z})}.
\label{eq12}
\end{equation}

(\romannumeral4) Apply the non-maxima suppression for each pixel,  the gradient modulus $\tilde{\xi}(\mathbf{z})$ and the gradient orientation $\theta(k)$ are used to decide whether it is a maximum of $\tilde{\xi}(\mathbf{z})$.

(\romannumeral5) Set the upper and lower thresholds, which are determined by the hisogram of the fused ESM of the input image. The size of the input image is $M\cdot N$, and $0$ $<$ $\beta_{low}$ $<$ $\beta_{up}$ $<$ $1$ are two percentiles specified in advance. The upper threshold $T_{up}$ and lower threshold $T_{low}$ are given by
\begin{equation}
T_{up}=\tilde{\xi}\left(\mathbf{z}_{\scriptscriptstyle[\beta_{up}\cdot M\cdot N]}\right),
\label{eq13}
\end{equation}

\begin{equation}
T_{low}=\tilde{\xi}\left(\mathbf{z}_{\scriptscriptstyle[\beta_{low}\cdot M\cdot N]}\right),
\label{eq14}
\end{equation}
where the symbol ``[]'' is zero-rounding operation, $\mathbf{z}_{\scriptscriptstyle[\beta_{up}\cdot M\cdot N]}$ and $\mathbf{z}_{\scriptscriptstyle[\beta_{low}\cdot M\cdot N]}$ represent the $[\beta_{up}\cdot M\cdot N]$th pixel and the $[\beta_{low}\cdot M\cdot N]$th pixel arranged by the value of ESM from small to large.

(\romannumeral6) Make the hysteresis determination. The decision of edge pixels is realized in two steps. All the pixels whose value of the fused ESM exceed $T_{up}$ are affirmed as edge pixels. If there is a path for the candidate edge pixels whose value of the fused ESM are between $T_{up}$ and $T_{low}$ to connect it with a strong edge pixel in the  four- or eight-neighborhood criterion, the candidate edge pixels are regarded as edge pixels.

\begin{figure}
\begin{center}
\begin{tabular}{c}
\includegraphics[width=15cm, height=7cm]{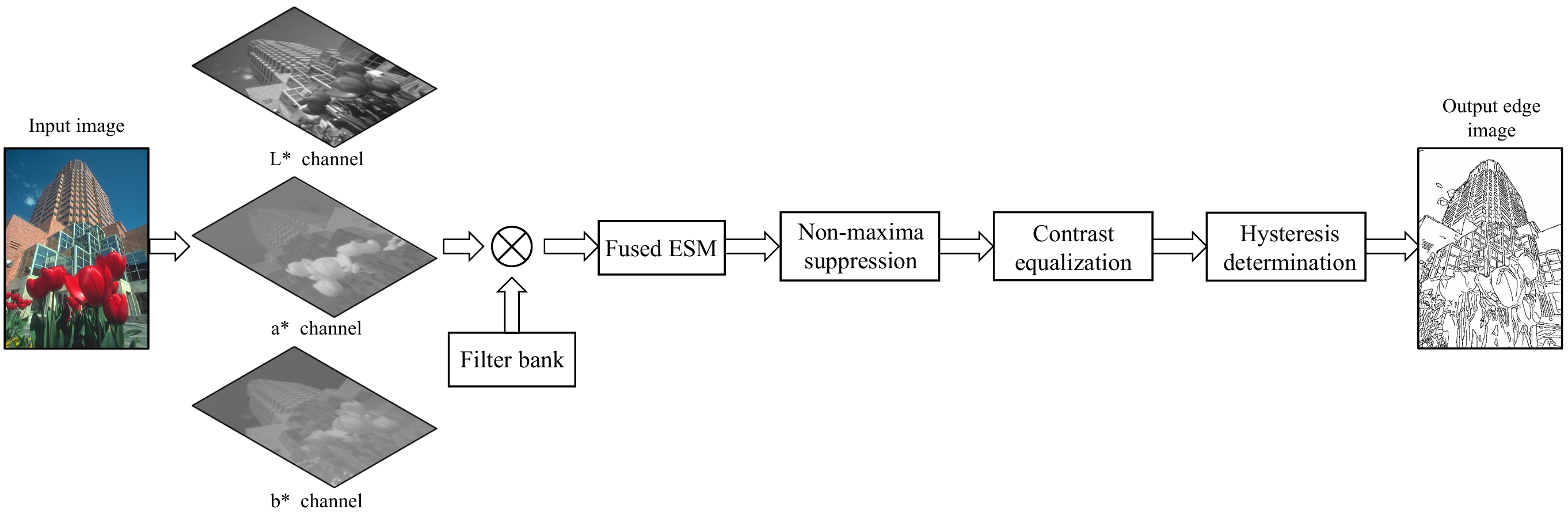}
\end{tabular}
\end{center}
\caption{The flowchart of framework of the proposed edge detection method.}
\label{f6}
\end{figure}

An example of the flow chart of the proposed algorithm is shown in Fig ~\ref{f6}.

\section{Experimental results and performance evaluation}
\label{sec:7}
This section reports the full performance evaluation results of the proposed edge detection method. The proposed method is compared with six detectors~\cite{evans2006a,Tai2008,Deng2011,akinlar2017colored,wang2016noise-robust,Kanade}. The criteria on precision-recall curve based on the Berkeley segmentation data set and benchmark 500 (BSDS500)~\cite{Arbelaez2011}, detection accuracy, and noise robustness are used to compare the performance of the five edge detection methods.

\subsection{PR curve assessment}
\label{sec:8}
The PR curve~\cite{Coleman2010,Dollr2014,Bae2015} is a recognized performance indicator for evaluating edge detection on image databases with ground truth maps (GTs). Precision is the ratio of the number of detected real edge pixels and the number of detected edge pixels. Recall is the ratio of the the number of detected real edge pixels and the number of edge pixels in the ground truth map.  For each detected edge pixel, the detected edge pixel will be marked as a true positive detected pixel if it matches the edge pixel in the ground truth edge map within a spatial tolerance distance. Otherwise, it will be marked as a false positive detection pixel. On the other hand, for a true edge pixel in ground truth edge map, if it is matched by the detected edge pixel within the spatial tolerance distance, the edge pixel in the ground truth map is marked as a matched ground truth pixel. Otherwise, it is treated as an unmatched benchmark pixel. In this way, the rates of precision and recall are calculated by
\begin{equation}
Y_{precision}=\frac{n_{TP}}{n_{TP}+n_{FP}},
\label{eq15}
\end{equation}

\begin{equation}
Y_{recall}=\frac{n_{MT}}{n_{MT}+n_{UM}},
\label{eq16}
\end{equation}
where $n_{TP}$, $n_{FP}$, $n_{MT}$, and $n_{UM}$ are true positive detection pixels, false positive detection pixels, matched ground truth pixels, and unmatched ground truth pixels respectively. In order to remove the influence of the parameter setup on evaluation, the parameter setup of the detector alters in an admissible parameter space so as to yield a large number of detected results and points on the plane of precision versus recall. For PR curve, there are two choices to set this threshold. The first one is referred as optimal dataset scale (ODS) which employs a fixed threshold for all images in a dataset. The second is called optimal image scale (OIS) which selects an optimal scale for each image. Three factors can be used to assess whether an algorithm is good or not : PR curve with a distance from the origin or area under the PR curve. It is worth to note that the area enclosed by the curve can also be evaluated by the parameter of average precision (AP)~\cite{Martin2004}. And the R50 is used to evaluate the recall when precision is
at $50\%$, which indicates the detection accuracy in high recall regime~\cite{Dollr2014}. The F-measure are calculated in this experiment.
\begin{equation}
 F=\frac{2\cdot Y_{precision}\cdot Y_{recall}}{Y_{precision}+Y_{recall}}.
 \label{eq17}
\end{equation}

\subsection{FOM index assessment}
\label{sec:9}
In this paper, Pratt's figure of merit (FOM)~\cite{pratt2001digital} is used to test the performance of the noise robustness of the five edge
detection methods. The missed true edge pixels, false detected edge pixels, and localization errors of detected edge pixels are considered to the calculation. Let the $n_{g}$ and the $n_{e}$  are sets of the numbers of the edge pixels in the ground truth map and the detected edge map respectively. Then, the FOM is defined as
\begin{equation}
 FOM=\frac{1}{max(n_{g},n_{e})}\sum_{j=1}^{n_{e}}\frac{1}{1+0.25d^2(j)},
 \label{eq18}
\end{equation}
where d(j) represents the distance from the jth detected edge pixel to the ideal edge map. $FOM=1$ indicates a perfect result.

\subsection{Experiment results based on the BSDS500 dataset}
\label{sec:10}
The BSDS500 is a widely used dataset in performance comparison on edge detection methods. It includes $200$ training, $100$ validation, and 200 test images. Each image is labeled by $4$ to $9$ annotators.

\begin{figure}
\begin{center}
\begin{tabular}{c}
\includegraphics[width=10cm, height=10cm]{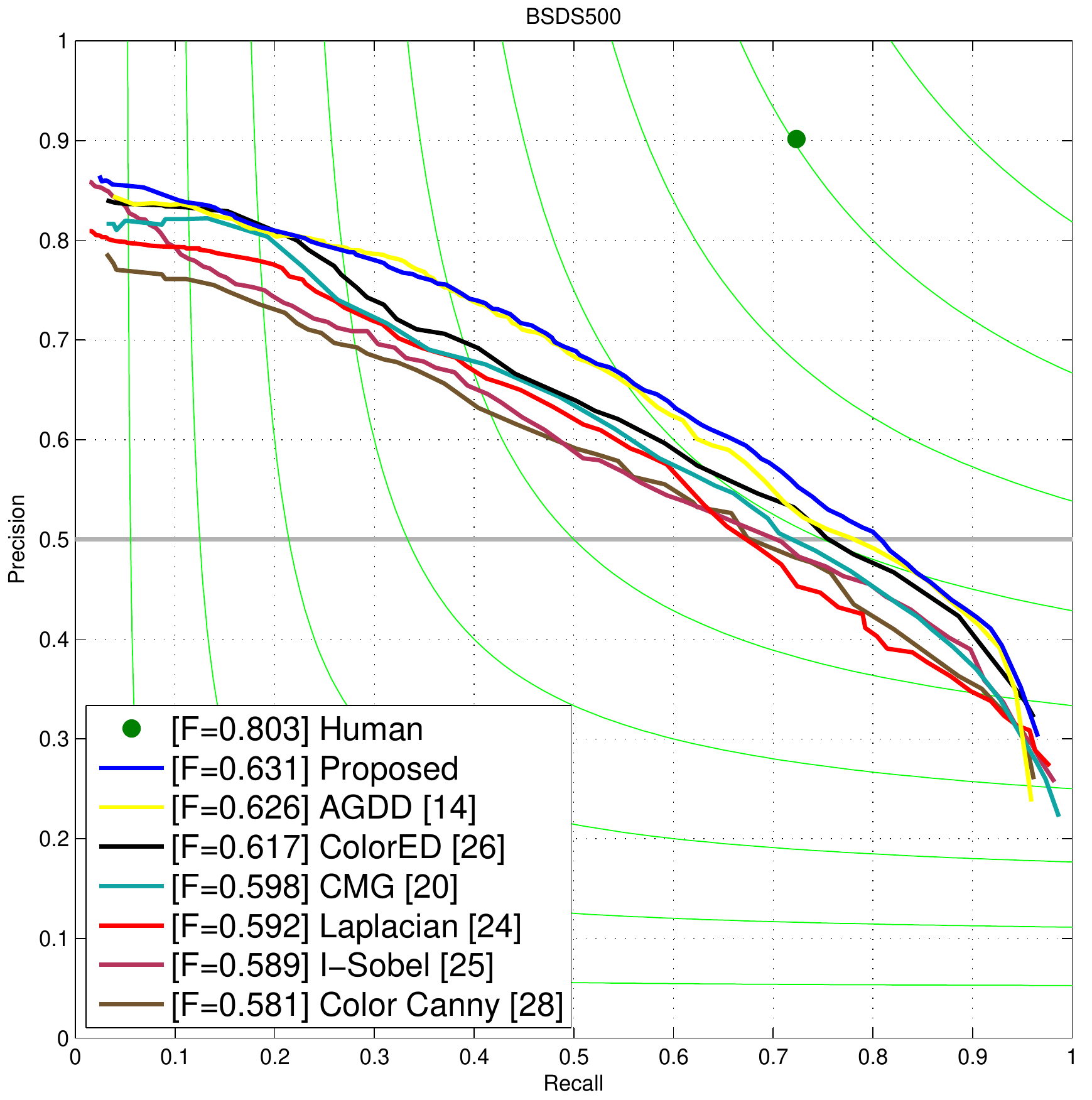}
\end{tabular}
\caption{PR curves of the five methods based on the BSDS500 dataset.}
\label{f7}
\end{center}
\end{figure}

The PR curves of the four edge detection methods are shown in Fig ~\ref{f7}. It can be observed that the proposed method achieves the maximum area under the PR curve. The reason is that the proposed method utilized the L*a*b* color space and the Gabor filters which have the ability to accurately obtain the local structure information from the input image. Furthermore, the proposed method applied the multi-scale and the multi-orientation technique which made the proposed method to have the better performance in noise robustness. Meanwhile, it can be found from Table~\ref{tab1} that the proposed method achieves better performance in terms of $F_{ODS}$, $F_{OIS}$, and $R50$ indexes.

\begin{table}
\centering
\caption{Evaluation metrics obtained by the seven methods on the BSDS500 dataset.}
\label{table}
\setlength{\tabcolsep}{10pt}
\begin{tabular}{ccccccc}
\hline
Methods&
$F_{ODS}$&
$F_{OIS}$&
AP&
R50\\
\hline
Color Canny~\cite{Kanade}&
 0.581 & 0.586 & 0.570 & 0.677\\
CMG~\cite{evans2006a}&
 0.598 & 0.605 & 0.607 & 0.719\\
Laplacian method~\cite{Tai2008}&
0.592 & 0.587 & 0.586 & 0.673\\
I-Sobel method~\cite{Deng2011}&
0.589 & 0.590 & 0.583 & 0.703\\
ColorED~\cite{akinlar2017colored}&
 0.617 & 0.619 & 0.618 & 0.756\\
AGDD~\cite{wang2016noise-robust}&
 0.626 & 0.627 & 0.632 & 0.781\\
Proposed&
 0.631 & 0.632 & 0.636 & 0.774\\
\hline
\end{tabular}
\label{tab1}
\end{table}

\begin{figure}
\begin{center}
\begin{tabular}{c}
\includegraphics[width=15cm, height=8cm]{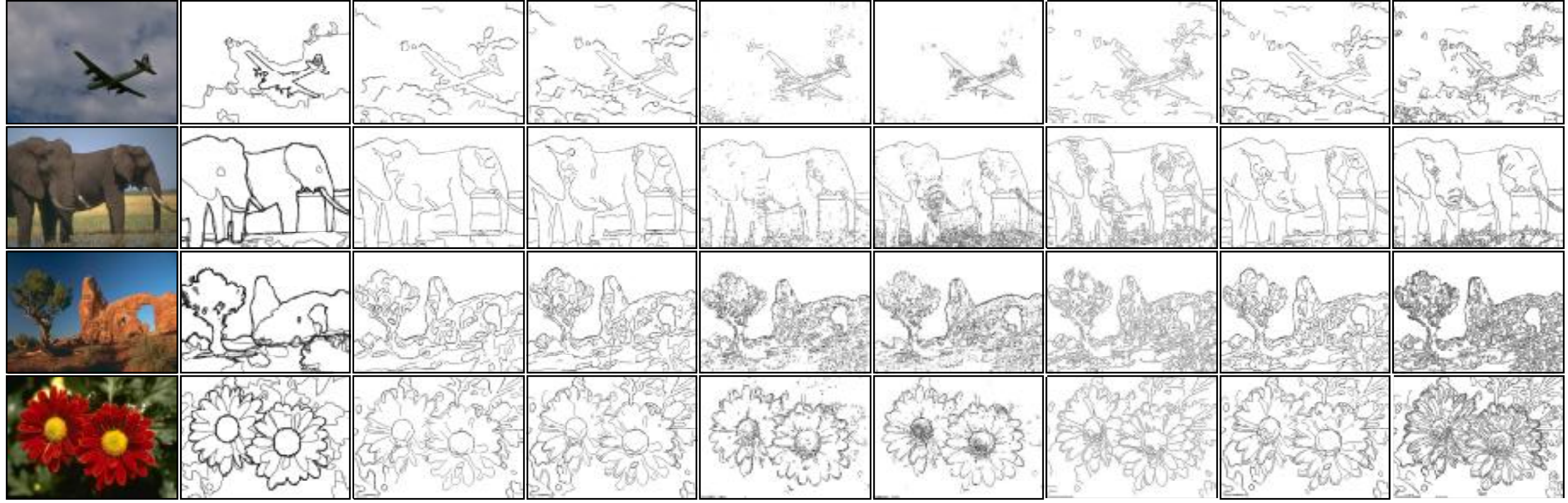}
\end{tabular}
\end{center}
\caption{Detection results of four methods on four test images.}
\label{f8}
\end{figure}

In this experiment, four original color images are shown in the first column of Fig ~\ref{f8}. The ground truths of the four original color images are shown in the second column of Fig ~\ref{f8}. The detection results of Color Canny method~\cite{Kanade}, CMG method~\cite{evans2006a}, Laplacian method~\cite{Tai2008}, I-Sobel method~\cite{Deng2011}, the AGDD method~\cite{wang2016noise-robust}, the ColorED method~\cite{akinlar2017colored} and the proposed method are shown in the third, fourth, fifth, sixth, seventh, eight and ninth columns of Fig ~\ref{f8} respectively. It can be observed that the proposed method can accurately detect edges from input images.

\begin{figure}
\begin{center}
\begin{tabular}{c}
\includegraphics[width=15cm, height=8cm]{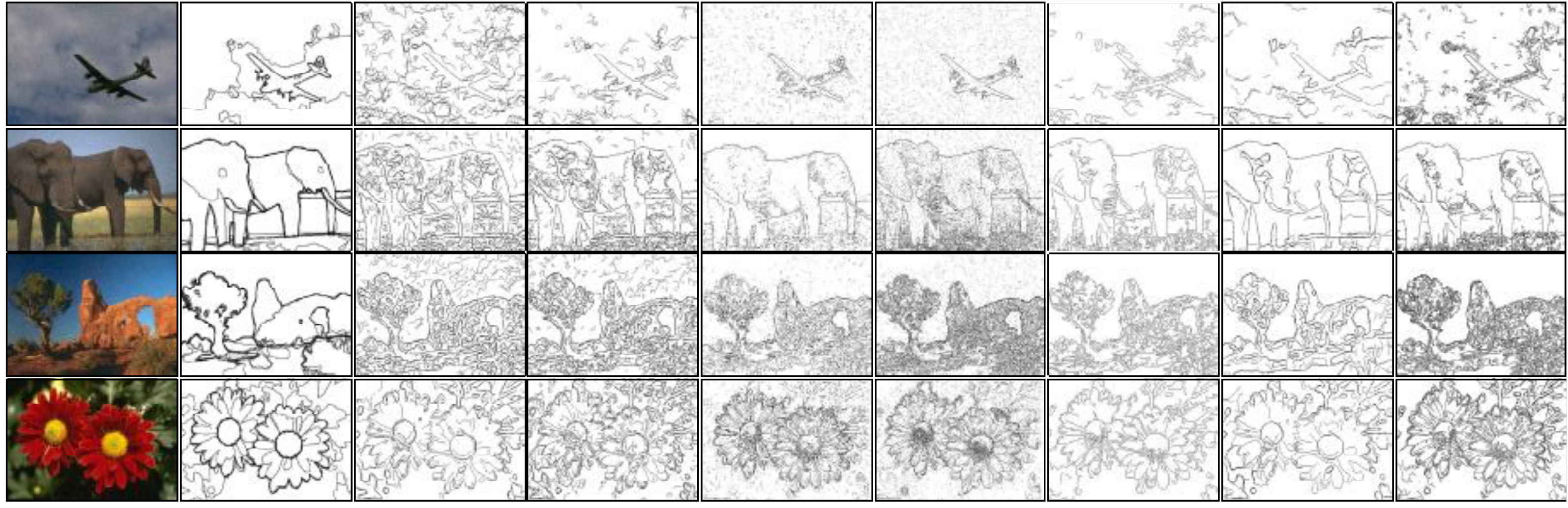}
\end{tabular}
\end{center}
\caption{Detection results of four methods on four noisy color images.}
\label{f9}
\end{figure}

Furthermore, the performance of the four edge detection methods are evaluated in noisy images with Gaussian standard deviation $\varepsilon_{w} = 15$. The noisy color images are shown in the first column of Fig ~\ref{f9}. The ground truth maps of the noisy color images are derived in terms of the FOM criteria~\cite{pratt2001digital} which are shown in the second column of Fig ~\ref{f9}. The detection results of Color Canny method~\cite{Kanade}, CMG method~\cite{evans2006a}, Laplacian method~\cite{Tai2008}, I-Sobel method~\cite{Deng2011}, the AGDD method~\cite{wang2016noise-robust}, the ColorED method~\cite{akinlar2017colored} and the proposed method are shown in the third, fourth, fifth, sixth, seventh, eighth and ninth columns of Fig ~\ref{f9} respectively. The FOMs of the five edge detection methods are summarized in Table ~\ref{tab2}. It can be observed that the proposed method has better noise
robustness.

\begin{table}[ht]
\centering
\caption{FOMs obtained by seven edge detection methods on the BSDS500 dataset.}
\label{table}
\setlength{\tabcolsep}{10pt}
\begin{tabular}{ccccccc}
\hline
Methods&
Plane&
Elephant&
Tree&
Flower\\
\hline
Color Canny~\cite{Kanade}&
 0.6675 & 0.6726 & 0.7731 & 0.7886  \\
CMG~\cite{evans2006a}&
 0.7223 & 0.7434 & 0.7745 & 0.7678  \\
Laplacian method~\cite{Tai2008}&
0.6127 & 0.6439 & 0.6328 & 0.6456\\
I-Sobel method~\cite{Deng2011}&
0.6517 & 0.6219 & 0.6473 & 0.6756\\
ColorED\cite{akinlar2017colored}&
 0.7713 & 0.7616 & 0.7925 & 0.8053  \\
AGDD~\cite{wang2016noise-robust}&
 0.7753 & 0.7727 & 0.8006 & 0.8124\\
Proposed&
 0.7837 & 0.7842 & 0.8075 & 0.8168 \\
\hline
\end{tabular}
\label{tab2}
\end{table}

\section{Conclusion}
\label{sec:11}
This paper proposed an edge detector with high edge detection accuracy and good noise robustness for color images. It used multi-directional Gabor filters with multiple scales to smooth the input image, which ia converted to the CIE L*a*b space. The fused ESM is from the ESMS of each channel at different scales. The proposed edge detection method is compared with five state-of-the-art edge detectors. Precision-recall curve based on the BSD500 dataset, edge detection accuracy, and noise robustness are used to assess the performance of the proposed edge detection method. The experimental results show that the proposed method is of very high quality.

\section{acknowledgements}
\label{sec:12}
This work was supported in part by the National Natural Science Foundation of China (NO. 61401347) and the Natural Science Basic Research Key Program funded by Shaanxi Provincial Science and Technology Department (2022JZ-35).

\bibliography{References}   
\bibliographystyle{unsrt}

\end{document}